\documentclass[letterpaper]{article} 
\usepackage{aaai2026}  
\usepackage{times}  
\usepackage{helvet}  
\usepackage{courier}  
\usepackage[hyphens]{url}  
\usepackage{graphicx} 
\usepackage{natbib}  
\usepackage{caption} 
\usepackage{algorithm}
\usepackage{algorithmic}

\usepackage{newfloat}
\usepackage{listings}

\usepackage{amsmath}
\usepackage{amssymb}
\usepackage{booktabs}
\usepackage{multirow}
\usepackage{tabularx}
\usepackage{xcolor}
\DeclareCaptionStyle{ruled}{labelfont=normalfont,labelsep=colon,strut=off} 
\floatstyle{ruled}
\newfloat{listing}{tb}{lst}{}
\floatname{listing}{Listing}
\title{SERL: Self-Examining Reinforcement Learning on Open-Domain}
\author{
    Weixuan Ou\textsuperscript{\rm 1}\equalcontrib,
    Yanzhao Zheng\textsuperscript{\rm 2}\equalcontrib,
    Shuoshuo Sun\textsuperscript{\rm 2},
    Wei Zhang\textsuperscript{\rm 2},
    Baohua Dong\textsuperscript{\rm 2}\thanks{Corresponding author},
    Hangcheng Zhu\textsuperscript{\rm 2},
    Ruohui Huang\textsuperscript{\rm 2},
    Gang Yu\textsuperscript{\rm 2},
    Pengwei Yan\textsuperscript{\rm 1},
    Yifan Qiao\textsuperscript{\rm 2}
}
\affiliations{
    \textsuperscript{\rm 1}Zhejiang University, Hangzhou, China\\

    \textsuperscript{\rm 2}Alibaba Group, Hangzhou, China\\
    \{ouweixuan, yanpw, pureidea.f\}@zju.edu.cn@stu.xjtu.edu.cn\\
    \{zhengyanzhao.zyz, sunshuoshuo.sss, zw443021, baohua.dbh, linran.lr09, wentong, ruohai\}@alibaba-inc.com\\

}

\usepackage{bibentry}

\begin{document}

\maketitle

\begin{abstract}
Reinforcement Learning (RL) has been shown to improve the capabilities of large language models (LLMs). However, applying RL to open-domain tasks faces two key challenges: (1) the inherent subjectivity of these tasks prevents the verifiable rewards as required by Reinforcement Learning with Verifiable Rewards (RLVR); (2) Reinforcement Learning from Human Feedback (RLHF) relies on external reward mechanisms. To overcome these limitations, we propose \textbf{S}elf-\textbf{E}xamining \textbf{R}einforcement \textbf{L}earning (\textbf{SERL}), a novel self-improving framework where the LLM serves as both Actor and Judge. SERL introduces two synergistic reward mechanisms without any external signals. On the one hand, to improve the Actor's capability, we derive rewards from Copeland-style pairwise comparison judgments across a group of generated responses. On the other hand, a self-consistency reward that encourages coherent judgments is proposed to enhance the Judge's reliability. This refinement strengthens the Judge, consequently generating a more robust training signal for the Actor. Experiments show that our method outperforms existing self-improvement training methods. SERL improves the LC win rate of Qwen3-8B on AlpacaEval 2.0 from 52.37\% to \textbf{59.90\%}. To the best of our knowledge, our method achieves state-of-the-art performance among self-improving approaches. Furthermore, it achieves a performance comparable to significantly larger models like Qwen3-32B, demonstrating superior effectiveness and robustness on open-domain tasks.
\end{abstract}

\begin{links}
    \link{Code}{https://github.com/AlwaysOu/SERL}
\end{links}

\begin{figure*}[t]
    \centering
    \includegraphics[width=0.95\textwidth]{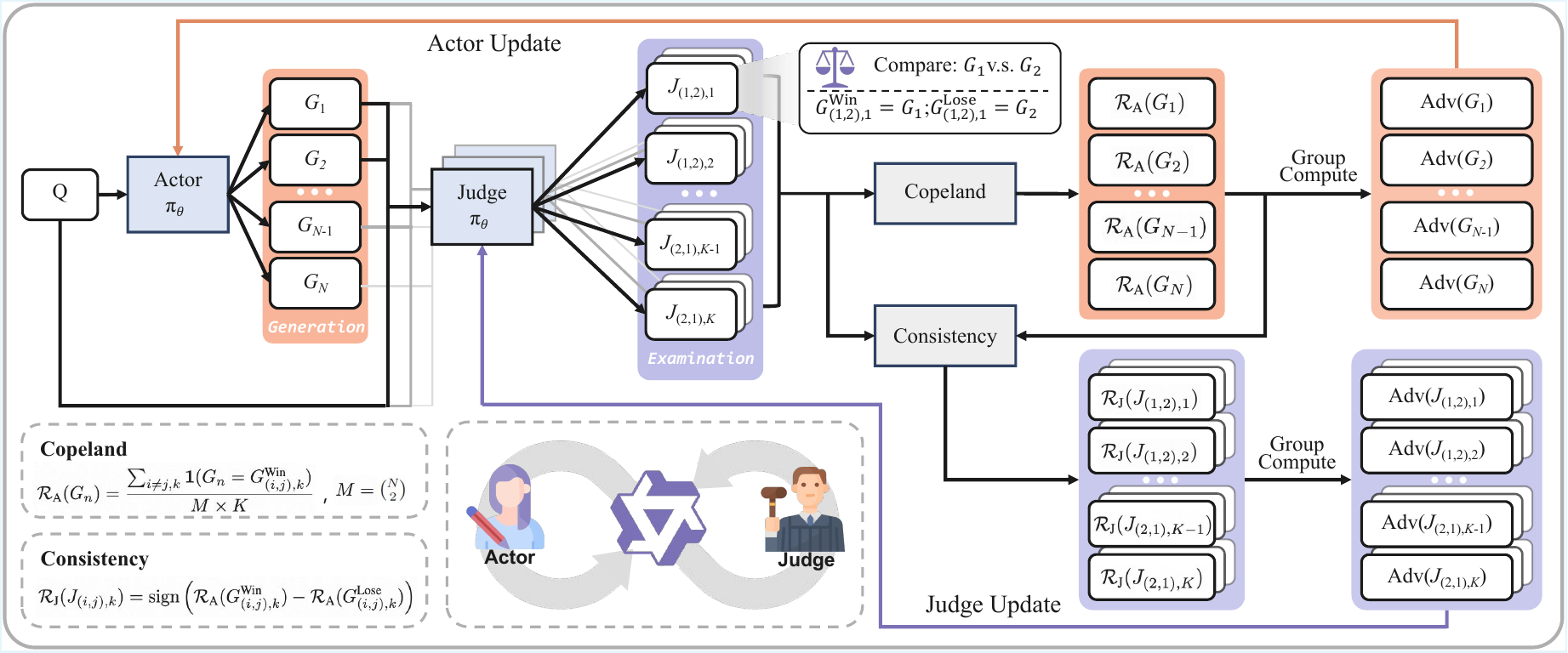}
    \caption{Overview of SERL. Given a instruction, the Actor first samples a group of responses. The Judge samples pairwise comparison judgments between response pairs. The judgments are aggregated via Copeland method to yield the Reward for Actor. Next, the consistency between the judgments and the ranking implied by the Reward for Actor is computed to generate the Reward for Judge. This process jointly enhances generation ability and comparative evaluation ability.}
    \label{fig:main_figure}
\end{figure*}

\begin{figure}[t]
    \centering
    \includegraphics[width=0.9\columnwidth]{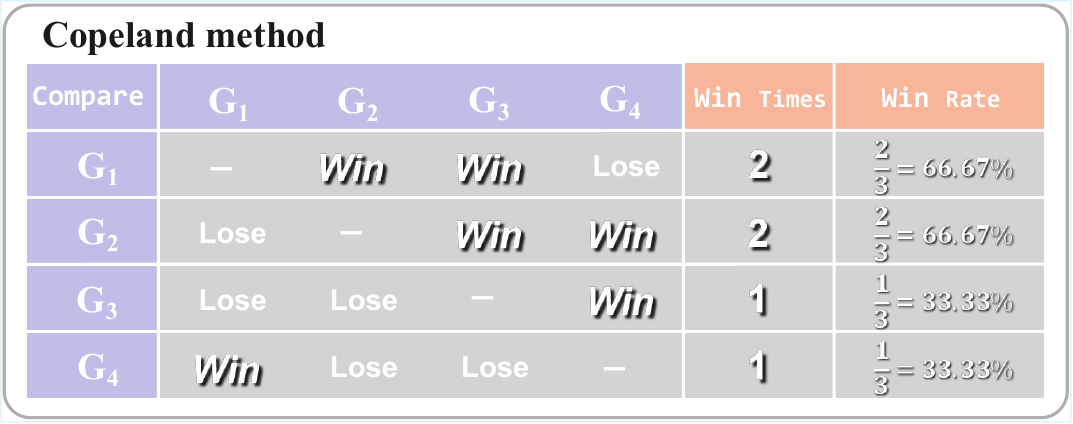}
    \caption{Illustration of the Copeland method.}
    \label{fig:copeland}
\end{figure}

\section{Introduction}
Reinforcement Learning (RL) has been demonstrated as an effective post-training approach that significantly improves the generation and reasoning capabilities of large language models~\cite{jaech2024openai, guo2025deepseek}. In particular, Reinforcement Learning with Verifiable Rewards (RLVR) has shown remarkable efficacy in specialized tasks such as mathematical reasoning~\cite{hu2025open, yu2025dapo} and code generation~\cite{luo2025deepcoder}. However, improving model performance on a broad range of open-domain tasks through RL still faces two major challenges.

\textbf{Unverifiable answers:} Open-domain tasks, including open writing and summarization, typically lack definitive correct answers, which are indispensable factors for RLVR.~\citet{ma2025general} try to solve this chanllenge by training a verification model to evaluate the consistency between policy answers and reference answers, while~\cite{zhou2025reinforcing, yu2025rlpr} leverage the policy itself to compute the joint distribution between the reasoning process and reference answers as rewards. These works only focus on objective question-answering tasks such as MMLU~\cite{wang2024mmlu}, GPQA~\cite{rein2024gpqa}, TheoremQA~\cite{chen2023theoremqa}, etc., while neglecting more open-ended task types such as summarization, open writing, and even general-domain open-QA. In practice, these tasks usually lack reference answers or supply only low-quality ones.

\textbf{Reliance on external reward mechanisms:} Earlier methodologies, such as Reinforcement Learning from Human Feedback (RLHF)~\cite{stiennon2020learning} and Reinforcement Learning from AI Feedback (RLAIF)~\cite{lee2023rlaif}, have demonstrated effectiveness in improving model capabilities across general domains by leveraging feedback from human annotators or AI-based evaluators. However, these RL methods also face significant limitations: they require extensive data annotation or dedicated reward models, leading to scalability challenges and additional computational overhead~\cite{gao2023scaling}.~\citet{yuan2024self,wu2024meta} introduce an offline self-improving method, in which the model scores each of its own outputs individually and then constructs preference data for DPO training in each iteration. These approaches require supervised cold-start fine-tuning, and their point-wise scoring depends on well-crafted standards, limiting their generalizability across diverse tasks.

To address these limitations, we propose SERL, a novel self-examining reinforcement learning framework designed specifically to enhance LLM generation capabilities in open-domain scenarios without relying on any external supervision signals or reward mechanisms. Our framework introduces a self-examining mechanism in which the model alternately assumes the roles of Actor and Judge, jointly optimizing its generation and evaluation capabilities. During training, the model samples diverse responses for each input and then evaluates these responses by itself. Specifically, we introduce Copeland-style judgments as shown in Figure \ref{fig:copeland}, in which model conducts pairwise comparisons between responses and subsequently ranks them according to their respective win rates within one group, establishing ranking rewards for generations and intrinsic consistency rewards for evaluations. SERL leverages the model's intrinsic evaluation capabilities while ensuring co-evolution of both generation and evaluation abilities via online learning, thus removing reliance on external reward signals.

Experiments on multiple open-domain tasks demonstrate the efficacy of SERL. Based on Qwen3-8B, our model achieves a notable performance increase on diverse open-domain tasks. Only a few dozen steps of training elevate the LC win rate on AlpacaEval 2.0 from 52.37\% to 59.90\%. Furthermore, extensive experiments across summarization and open writing tasks show that SERL consistently outperforms other mainstream self-improving methods, and delivering performance competitive with significantly large models like Qwen3-32B.

In summary, our main contributions are as follows:

\begin{enumerate}
\item We introduce SERL, a novel self-examining reinforcement learning framework that enables large language models to jointly optimize their generation and evaluation abilities without any external supervision signals, verifier models, or human annotations, thereby inherently enjoying scalability and flexibility on open-domain tasks.

\item Extensive experiments show that SERL achieves remarkable training efficiency and performance improvements on diverse open-domain tasks, including summarization, open writing, and General-QA. With only dozens of training steps, SERL boosts Qwen3-8B’s LC win rate on AlpacaEval 2.0 by 7.53\%, and delivers competitive or superior performance to much larger models such as Qwen3-32B. To the best of our knowledge, our method achieves state-of-the-art performance among self-improving approaches.
\end{enumerate}

\section{Related Works}
\subsubsection{RLVR}
Works on Reinforcement Learning with Verifiable Rewards (RLVR) have achieved improvements in the capabilities of large models. Representative works include CodeRL~\cite{le2022coderl}, RLTF~\cite{liu2023rltf}, and PPOCoder~\cite{shojaee2023execution} in programming.  
Recently, Group Relative Policy Optimization (GRPO)~\cite{shao2024deepseekmath} improves sample efficiency by estimating advantages within a sampled group without a learned value network. GRPO and follow-up works such as DAPO~\cite{yu2025dapo} and VAPO~\cite{yue2025vapo} demonstrate continued improvements in math. However, they cannot be directly applied to open-domain settings due to the lack of verifiable answers for these tasks.
SERL preserves GRPO’s group-wise efficiency while computing differential rewards through pairwise comparison judgments, thereby overcoming the reward-design bottleneck on open-domain tasks where verifiable rewards are unavailable.

\subsubsection{RLHF and RLAIF}
Reinforcement Learning from Human Feedback (RLHF) involves training a reward model on human pairwise comparisons and subsequently optimizing the policy with algorithms like PPO. This approach has achieved strong alignment in prominent models such as InstructGPT~\cite{ouyang2022training}, ChatGPT/GPT-4~\cite{achiam2023gpt}, and Llama-2-chat~\cite{touvron2023llama}. To improve efficiency, variants like Direct Preference Optimization (DPO)~\cite{rafailov2023direct}, KTO~\cite{ethayarajh2024kto}, and Online-DPO~\cite{lanchantin2025bridging} simplify or accelerate preference learning, yet still depend on human labels.
Reinforcement Learning from AI Feedback (RLAIF)~\cite{bai2022training} was introduced to alleviate the costly process of human annotation, which uses the LLM itself as the labeler. This includes methods like Constitutional AI~\cite{bai2022constitutional}, which apply fixed principles for self-critique. As noted in Self-Rewarding~\cite{yuan2024self}, employing the Actor as a Judge to provide feedback has been shown to enhance the Actor's capabilities. However, it lacks optimization for Judge. In contrast, Meta-Rewarding~\cite{wu2024meta} jointly optimize both the Actor and the Judge, but it relies on off-policy learning paradigms. 
We propose SERL, an on-policy training framework that jointly optimizes the model's capabilities as both Actor and Judge, independent of external signals. In SERL, the same model simultaneously acts as Actor to generate multiple reasoning traces and as Judge to compare them, thereby obviating the need for external human or proxy preference labels.

\section{Method}
SERL is a self-improving framework designed to comprehensively improve a model's performance on open-domain tasks. Its methodology is centered on the model assuming two primary roles: Actor and Judge. As the Actor, it generates responses to given prompts. As the Judge, it compares the relative strengths and weaknesses among its own generated responses. The rewards for Actor and Judge are constructed without reliance on any external signals, synergistically improving the model's generation and evaluation capability.

SERL's training pipeline (Figure \ref{fig:main_figure}) is as follows:
\begin{itemize}
    \item \textbf{Generation.}
    For each prompt, the Actor stochastically samples a set of diverse responses. This step creates the raw material for examination and refinement.
    \item \textbf{Examination.}
    For each response pair, the Judge employs an ``LLM-as-a-Judge'' prompting strategy to sample a set of pairwise comparison judgments. 
    \item \textbf{Rewards.}
    For a set of pairwise comparison judgments, Reward for Actor ($\mathcal{R}_\text{A}$) is calculated as the win rate of each response across all pairwise comparisons. $\mathcal{R}_\text{A}$ for each response can be interpreted as its rank among the generated responses. Reward for Judge ($\mathcal{R}_\text{J}$) is calculated as the consistency between this ranking and each individual pairwise comparison judgment. 
\end{itemize}

This dual-reward mechanism is analogous to how humans approach complex problems by attempting multiple solutions, comparing the effectiveness of these approaches, and ultimately becoming proficient in solving such problems. Without relying on any external signals, this mechanism refines model's evaluation capability (as Judge), which in turn provides a more robust reward to enhance model's generation capability (as Actor).

\subsection{Generation}

At step \(t\), given a set of questions \(Q= \{q\}\), we sample a group of N individual responses for each input \(q\) from the old Actor \(\pi_{\mathrm{old}}\): \(\{ G_n \}^{N}_{n=1} \sim \pi_{\mathrm{old}}\).

\subsection{Examination}
We deliberately choose comparison as the core evaluation mechanism because of the widely recognized principle that for both humans and LLMs~\cite{ouyang2022training,zheng2023judging}, making relative judgments (i.e., ``Is response A better than response B?'') is a more tractable and reliable task than assigning absolute scores (e.g., ``Rate response A on a scale of 1 to 10''). For open-domain tasks that lack verifiable answers and explicitly defined evaluation criteria, framing the evaluation as a series of forced-choice preference selections enables us to elicit more stable, consistent, and meaningful feedback, which is crucial for effectively guiding the iterative refinement of our models.

To aggregate these individual preference judgments into a coherent global ranking, we employ the Copeland method~\cite{copeland1951reasonable,dwork2001rank}. As illustrated in Figure \ref{fig:copeland}, this method treats each response as a ``candidate'' and tallies the results of all possible pairwise matchups. This approach is not only intuitive and robust but also effectively resolves potential preference cycles, providing a clear and defensible final ordering.

At step $t$, the Judge evaluates comparisons to determine the preferred response between $G_i$ and $G_j$, where $i,j \in \{1,...,N\},~i \neq j$. To enhance the reliability of this process, we employ a multi-sample evaluation strategy. For each pair $(G_i, G_j)$, we sample $K$ independent judgments $J_{(i,j),k}$ from the old Judge \(\pi_{\mathrm{old}}\): \(\{ J_{(i,j),k} \} \sim \pi_{\mathrm{old}}\). We define $G_{(i,j),k}^{\text{Win}}$ and $G_{(i,j),k}^{\text{Lose}}$ as the winning and losing responses, in the comparison between $\{G_i, G_j\}$, based on the $k$-th judgment $J_{(i,j),k}$.

\subsubsection{Position Bias Mitigation Mechanism}
Some works have found that ``LLM-as-a-Judge'' exhibits position bias, the tendency of judgments to disproportionately favor responses based on their placement within an input list rather than their intrinsic merit~\cite{shi2024judging}. To mitigate positional bias, we introduce the Position Bias Mitigation Mechanism (PBMM) that swaps the positions in half of the \( K \) pairwise comparisons: $K//2$ prompts constructed as \((q, G_i, G_j)\), and $K//2$ constructed as \((q, G_j, G_i)\). 

\subsection{Rewards}
\subsubsection{Reward for Actor}  
We introduce the metric $\mathcal{R}_\text{A}$ to enhance model's generation capability. Specifically, $\mathcal{R}_\text{A}$ for each response \( G_n \) is defined as its win rate across all pairwise comparisons with other candidate responses. This metric provides a clear, direct quantification of the relative quality of \( G_n \) compared to other responses, thus offering an explicit and informative learning signal to guide and optimize the Actor's generation behavior. $\mathcal{R}_\text{A}$ is calculated as follows:
\begin{equation}
\label{eq:actor_reward}
\mathcal{R}_\text{A}(G_n) = \frac{\sum_{i \neq j, k} \mathbf{1}(G_n = G_{(i,j),k}^{\text{Win}})}{M \times K}
\end{equation}

where \(M\) is calculated as \(\binom{N}{2}\), the number of pairwise combinations involving \(\{ G_n \}\).

\subsubsection{Length Control Module}
Length-bias means the model's tendency to favor longer responses~\cite{dubois2024length,park2024disentangling}. To ensure the evaluations are conducted under similar output length conditions, we introduce a Length Control Module (LCM). Accordingly, the $\mathcal{R}_\text{A}$ is further modified as follows:

\begin{equation} \label{actor_reward_limit}
\mathcal{R}_\text{A}(G_n) = \frac{\sum_{i \neq j, k} \beta \left( G_n = G_{(i,j),k}^{\text{Win}}\right)}{M \times K}
\end{equation}

\begin{equation}
    \beta = \frac{
    |(G_{(i,j),k}^{\text{Lose}})|}{|(G_{(i,j),k}^{\text{Win}})|}
\end{equation}

\begin{equation}
\text{s.t.}\ (1-\alpha) < \frac{|(G_{(i,j),k}^{\text{Win}})|}{|(G_{(i,j),k}^{\text{Lose}})|} < (1+\alpha)
\notag
\end{equation}

The hyperparameter $\alpha$ is used to restrict valid comparisons to only those pairwise responses where lengths are sufficiently similar. Specifically, a smaller value of $\alpha$ imposes a stricter requirement for the lengths of the compared responses to be closer. In our setup, $\alpha=0.2$. 
The ratio $\beta$ serves to balance the influence of response length on the win-loss determination among the valid comparisons, assigning a greater reward to shorter winning responses.

\subsubsection{Reward for Judge}
$\mathcal{R}_\text{J}$ is proposed to improve model's evaluation capabilities. We define $\mathcal{R}_\text{J}$ to measure the consistency between the global ranking of responses and each pairwise comparison judgement. A judgment is considered consistent if the selected winner in a pairwise comparison ranks higher than the loser in the global ranking. Higher consistency indicates the model's evaluative confidence, offering clear signals for reliable and coherent judgment decisions. $\mathcal{R}_\text{J}$ is described as follows:

\begin{equation} \label{judge_reward_unified}
\mathcal{R}_\text{J}(J_{(i,j),k}) = \operatorname{sign}\left(\mathcal{R}_\text{A}(G_{(i,j),k}^{\text{Win}}) - \mathcal{R}_\text{A}(G_{(i,j),k}^{\text{Lose}})\right)
\end{equation}

where the sign function is defined as:

\begin{equation}
\operatorname{sign}(x) = \frac{x}{|x|} \quad (x \neq 0), \quad \operatorname{sign}(0) = 0
\end{equation}

\subsection{Online Optimization}
The model is optimized on a mixed set of responses and judgments. The SERL training objective, adapted from GRPO, is formulated with a surrogate function \(\text{Surr.}\) as:

\begin{equation}
\label{eq:grpo_style}
\begin{aligned}
\mathcal{J}_{\text{GRPO}}(\theta)
&= \mathbb{E}_{\substack{q\sim Q,\\ G\sim\pi_{\text{old}}}}
\!\Biggl[
      \frac{1}{N}
      \sum_{n=1}^{N}
      \text{Surr.}(\text{Actor},{G_n},{n};{\theta})\Biggr]
\end{aligned}
\end{equation}

\begin{equation}
\label{eq:serl_grpo_style}
\begin{aligned}
\mathcal{J}_{\text{SERL}}(\theta)
&= \mathbb{E}_{\substack{q\sim Q,\\ G,J\sim\pi_{\text{old}}}}
\!\Biggl[
      \frac{1}{N}
      \sum_{n=1}^{N}
      \text{Surr.}(\text{Actor},{G_n},{n};{\theta}) \\
&\hspace{-5em} \quad +
      \frac{1}{M \times K}
      \sum_{i\neq j}
      \sum_{k=1}^{K//2}
      \text{Surr.}(\text{Judge},{J_{(i,j),k}},{[(i,j),k]};{\theta})
\Biggr]
\end{aligned}
\end{equation}

where the objective function \(\text{Surr.}\) for a single sampled output, the advantages calculated by normalizing the group-level rewards \(\hat{A}^{\text{Actor}}_{n,t}\), \(\hat{A}^{\text{Judge}}_{(i,j),k,t}\) for all tokens in a response, and the token-level probability ratios \(r^{\text{Actor}}_{n,t}(\theta)\), \(r^{\text{Judge}}_{(i,j),k,t}(\theta)\), are defined as:
\begin{equation}
\begin{aligned}
&\text{Surr.}({R},{o},{i};~{\theta})
= \\
&\frac{1}{|o|}
\sum_{t=1}^{|o|}
\min\Biggl[
    r^{R}_{i,t}(\theta)\,
    \hat{A}^{R}_{i,t}, 
\operatorname{clip}\Bigl(
    r^{R}_{i,t},
    1 - \epsilon,\,
    1 + \epsilon
  \Bigr)\,
  \hat{A}^{R}_{i,t}
\Biggr]
\end{aligned}
\end{equation}

\begin{equation}
\hat{A}^{\text{Actor}}_{n,t} = \frac{\mathcal{R}_\text{A} - \mathrm{mean}(\{ \mathcal{R}_\text{A} \}^N)}{\mathrm{std}(\{ \mathcal{R}_\text{A} \}^N)}
\end{equation}

\begin{equation}
\hat{A}^{\text{Judge}}_{(i,j),k,t} = \frac{\mathcal{R}_\text{J} - \mathrm{mean}(\{ \mathcal{R}_\text{J} \}^{M \times K})}{\mathrm{std}(\{ \mathcal{R}_\text{J} \}^{M \times K})}
\end{equation}

\begin{equation}
r^{\text{Actor}}_{n,t}(\theta) = \frac{\pi_\theta(G_{n,t}\,|\,q, G_{n,<t})}{\pi_{\theta_{\mathrm{old}}}(G_{n,t}\,|\,q, G_{n,<t})}
\end{equation}

\begin{equation}
r^{\text{Judge}}_{(i,j),k,t}(\theta) = \frac{\pi_\theta(J_{(i,j),k,t}\,|\,q, J_{(i,j),k,<t})}{\pi_{\theta_{\mathrm{old}}}(J_{(i,j),k,t}\,|\,q, J_{(i,j),k,<t})}
\end{equation}

In GRPO, the KL penalty term is used to regulate the divergence between the online policy and the frozen reference policy. However, recent work suggests that during the training of long-CoT reasoning model, the model distribution can diverge significantly from the initial model~\cite{yu2025dapo}. Therefore, we exclude the KL term in SERL.

\section{Experiments}
\setcounter{table}{0} 

\begin{table*}[t]
\small
\centering
\begin{tabular}{ll ccc ccc}
\toprule
\multirow{2}{*}{\textbf{Method1}} & \multirow{2}{*}{\textbf{Method2}} 
  & \multicolumn{3}{c}{\textbf{Summarization}} 
  & \multicolumn{3}{c}{\textbf{Open Writing}} \\
\cmidrule(lr){3-5} \cmidrule(lr){6-8}
  & 
  & Win rate & $\Delta$Win rate & Len & Win rate & $\Delta$Win rate & Len\\
\midrule
SERL(Ours)  & Online-DPO    & 55.17\% & +10.33\% & 1190 : 1797 & 50.50\% & +1.00\% & 1654 : 1767 \\
SERL(Ours)  & Self-Rewarding     & 59.50\% & +19.00\% & 1190 : 922 & 55.17\% & +10.33\% & 1654 : 1887 \\
SERL(Ours)  & Meta-Rewarding       & 59.17\% & +18.33\% & 1190 : 1441 & 56.67\% & +13.33\% & 1654 : 2189\\
SERL(Ours)  & RLSC     & 86.17\% & +72.33\% & 1190 : 1322 & - & - & -\\
SERL(Ours)  & GRPO(ROUGE-L)       & 99.17\% & +98.33\% & 1190 : 248 & - & - & -\\
\bottomrule
\end{tabular}
\caption{Evaluation results of SERL against other methods on summarization and open writing tasks. $\Delta$Win rate is the difference in win rates between two methods.}
\label{tab:ex2.1}
\end{table*}

\begin{table}[ht]
\centering
\begin{tabular}{l ccc}
\toprule
\multirow{2}{*}{\textbf{Models}} 
  & \multicolumn{3}{c}{\textbf{General QA (AlpacaEval 2.0)}} \\
\cmidrule(lr){2-4}
  & LC win rate & Win rate & Len \\
\midrule
Online-DPO    & 54.07\%  & \underline{59.74\%} & 3429 \\
Self-Rewarding    & 51.29\%  & 53.69\% & 3074 \\
Meta-Rewarding  & \underline{54.73\%}  & 55.93\% & 3081 \\
RLSC  & 52.11\%  & 51.81\% & 2060 \\
SERL(Ours)   & \textbf{59.90\%}  & \textbf{69.88\%} & 3017 \\
\bottomrule
\end{tabular}
\caption{Evaluation results of SERL and other methods on general QA task.}
\label{tab:ex2.2}
\end{table}

\subsection{Experimental Setup}
\subsubsection{Datasets} To comprehensively evaluate the effectiveness and generalization capability of our proposed method SERL, we conduct experiments on three tasks including summarization, open writing, and general QA.

\begin{itemize}
\item \textbf{Summarization} CNN-DM~\cite{hermann2015teaching}: A standard dataset for abstractive text summarization, consisting of news articles and their corresponding human-written summaries.

\item \textbf{Open Writing} writingprompts~\cite{fan2018hierarchical}: A large-scale dataset containing open writing prompts and human-written stories.

\item \textbf{General QA}
(1) AlpacaEval 2.0~\cite{dubois2024length}: A recently developed benchmark that includes a wide range of general QA tasks spanning multiple domains. This allows us to test the general applicability of our method in handling diverse questions and producing accurate and coherent responses. (2) Ultrafeedback~\cite{ye2023ultrafeedback}: A large-scale, diverse, and fine-grained preference dataset. We select this dataset as the training dataset for the general QA task because it contains diverse prompts from datasets such as TruthfulQA, FalseQA, Evol-Instruct, UltraChat, and ShareGPT.
\end{itemize}

As SERL is independent of external reward signals, we use only the prompts from these datasets. 

\subsubsection{Baselines}
\begin{itemize}
\item \textbf{Self-Rewarding}~\cite{yuan2024self}
Employs the model itself as a reward estimator to iteratively refine its own outputs, but lacks formal optimization for the reward estimator.
\item \textbf{Meta-rewarding}~\cite{wu2024meta}
Jointly optimizes the Actor and Judge via off-policy learning, decoupling feedback generation from policy updates.
\item \textbf{Online-DPO}~\cite{lanchantin2025bridging}
Recent work has found that Online-DPO achieves performance comparable to GRPO. In self-improving setting, we adopt the policy model itself as a reward model to compare sampled pairwise responses.
\item \textbf{RLSC}~\cite{li2025confidence}
This method introduces confidence-aware reinforcement learning, also without reliance on any external signals, to balance exploration and exploitation, showing improvement in math.
\item \textbf{GRPO}~\cite{shao2024deepseekmath}
In summarization task, we employ the ROUGE-L score~\cite{lin2003automatic}, computed by comparing the output against the reference summary from the dataset, as the reward for GRPO.
\end{itemize}

To ensure a fair comparison, we remove the Supervised Fine-Tuning (SFT) on the Instruction Fine-Tuning (IFT) and Evaluation Fine-Tuning (EFT) data for the Self-Rewarding and Meta-Rewarding methods, which would introduce an external training signal.

\subsubsection{Implementation Details and Evaluation Metrics}
We adopt the Qwen3-8B as base model for all methods. All baselines use their authors’ recommended hyperparameters. For SERL, we set the sample number of Actor \(N\) as 4, the sample number of Judge \(K\) as 4, topP=1.0, topK=50, and temperature = 0.9 during generation and examination. Other parameters like epoch and learning rate are set differently for various tasks.

For both summarization and open writing tasks, we select GPT-4o as the evaluator to compare two outputs—either summaries or story continuations—generated from two methods on the same instruction, selecting the better output. To reduce positional bias, we compared the outputs from two methods twice. For the second comparison, we swapped the order of the two responses. If both pairwise comparisons identify the same model's output as the winner, then give a WIN for that model and a LOSE to the other; otherwise, the result is considered a TIE. For these two tasks, we define the win rate as follows: 

\begin{equation}
\label{eq:winrate}
\text{Win rate} = \frac{2 \times \text{WIN} + \text{TIE}}{2 \times \text{WIN} + \text{TIE} + 2 \times \text{LOSE}}
\tag{5}
\end{equation}

For the general QA task, we evaluate on AlpacaEval 2.0, reporting the win rate and Length-controlled (LC) win rate.

More details of implementation of SERL and Baselines can be found in Appendix\ref{app:Implementation of SERL and Baselines}.

\begin{table*}[t]
\small
\centering
\begin{tabular}{ll ccc ccc}
\toprule
\multirow{2}{*}{\textbf{Model1}} & \multirow{2}{*}{\textbf{Model2}} 
  & \multicolumn{3}{c}{\textbf{Summarization}} 
  & \multicolumn{3}{c}{\textbf{Open Writing}} \\
\cmidrule(lr){3-5} \cmidrule(lr){6-8}
  & 
  & Win rate & Win rate$\uparrow$ & Len & Win rate & Win rate$\uparrow$ & Len\\
\midrule
Qwen3-8B(base)  & Qwen3-32B    & 37.33\% & - & 1066 : 1109 & 33.00\% & - & 1369 : 1506\\
Qwen3-8B(base)  & R1-Distill-Qwen-32B     & 63.50\% & - & 1066 : 922 & 78.50\% & - & 1369 : 1754\\
Qwen3-8B(base)  & R1-Distill-Llama-70B     & 63.67\% & - & 1066 : 1012 & 71.33\% & - & 1369 : 2023\\
Qwen3-8B(base)  & Claude 3.5 Sonnet     & 48.33\% & - & 1066 : 1064 & 81.33\% & - & 1369 : 1632\\
Qwen3-8B(base)  & GPT-4o-0513       & 55.33\% & - & 1066 : 941 & 63.50\% & - & 1369 : 1412\\
\midrule
SERL(Ours)  & Qwen3-8B(base)    & 62.83\% & +12.83\% & 1190 : 1066 & 61.50\% & +11.50\% & 1654 : 1369\\
\midrule
SERL(Ours)  & Qwen3-32B   & 52.67\% & +15.33\% & 1190 : 1109 & 46.67\% & +13.67\% & 1654 : 1506\\
SERL(Ours)  & R1-Distill-Qwen-32B     & 72.17\% & +8.67\% & 1190 : 922 & 88.67\% & +10.17\% & 1654 : 1754\\
SERL(Ours)  & R1-Distill-Llama-70B     & 71.00\% & +7.33\% & 1190 : 1012 & 84.00\% & +12.67\% & 1654 : 2023\\
SERL(Ours)  & Claude 3.5 Sonnet     & 56.67\% & +8.33\% & 1190 : 1064 & 91.00\% & +9.67\% & 1654 : 1632\\
SERL(Ours)  & GPT-4o-0513       & 65.50\% & +10.17\% & 1190 : 941 & 73.67\% & +10.17\% & 1654 : 1412\\
\bottomrule
\end{tabular}
\caption{Evaluation results of SERL against other general-purpose LLMs on summarization and open writing tasks. Win rate$\uparrow$ is the relative win rate improvement, which is calculated as the difference between the win rate of the model trained with our method against other models and the win rate of the base model against the same models.}
\label{tab:ex1.1}
\end{table*}

\begin{table}[ht]
\small
\centering
\begin{tabular}{l ccc}
\toprule
\multirow{2}{*}{\textbf{Models}} 
  & \multicolumn{3}{c}{\textbf{General QA (AlpacaEval 2.0)}} \\
\cmidrule(lr){2-4}
  & LC win rate & Win rate & Len \\
\midrule
Qwen3-8B    & 52.37\%  & 55.07\% & 3100 \\
Qwen3-32B  & \textbf{62.16\%}  & \underline{66.47\%} & 3034 \\
R1-Distill-Qwen-32B   & 41.76\%  & 37.79\% & 1817 \\
R1-Distill-Llama-70B   & 54.06\%  & 44.51\% & 1617 \\
Claude 3.5 Sonnet     & 57.77\% & 44.70\% & 1487 \\
GPT-4o-0513   & 57.46\%  & 51.33\% & 1873 \\
SERL(Ours)   & \underline{59.90\%}  & \textbf{69.88\%} & 3017 \\
\bottomrule
\end{tabular}
\caption{Evaluation results of SERL and other general-purpose LLMs on general QA task.}
\label{tab:ex1.2}
\end{table}

\begin{figure}[t]
    \centering
    \includegraphics[width=0.95\columnwidth]{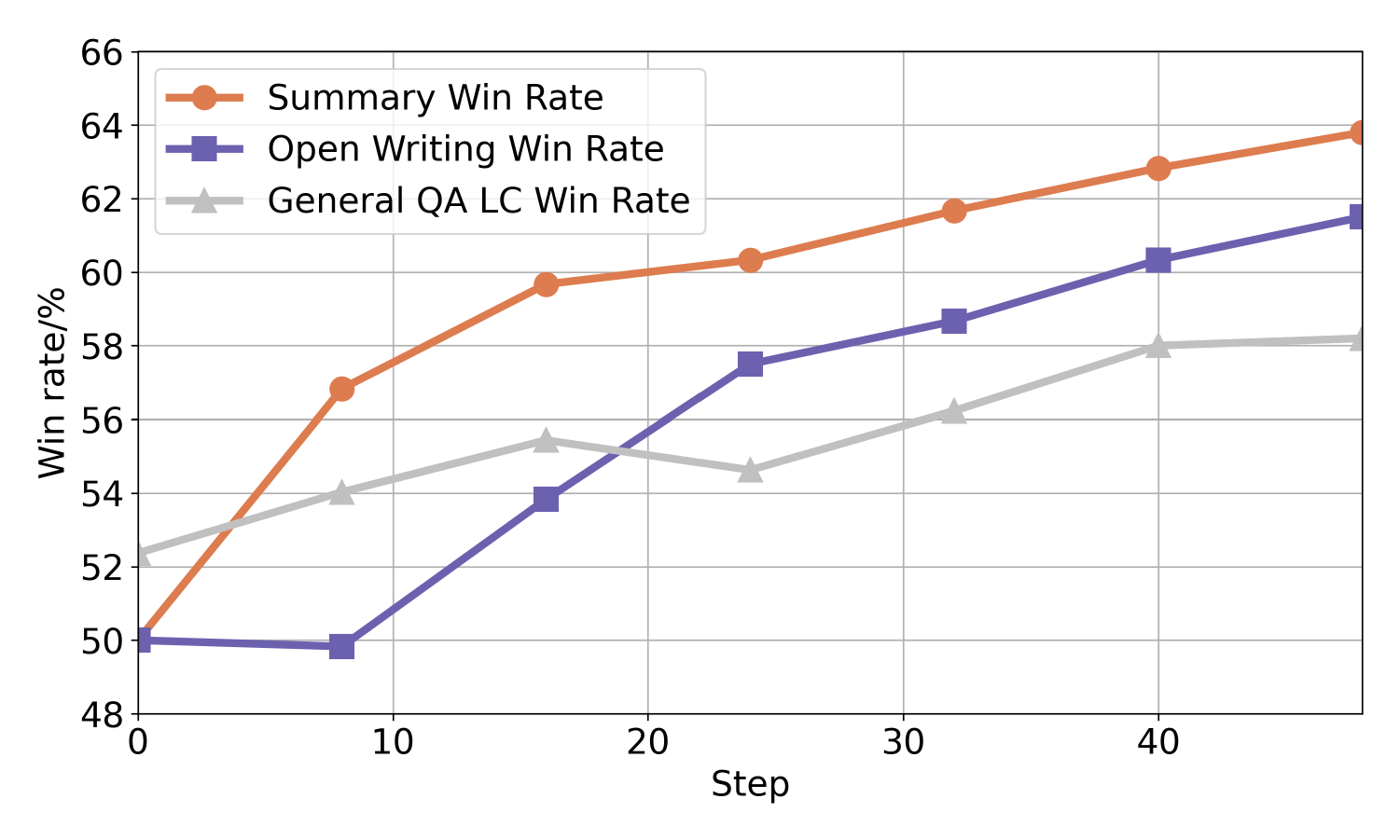}
    \caption{The win rate against Qwen3-8B on summarization and open writing, and the LC win rate on general QA with SERL training step.}
    \label{fig:win_rate_step}
\end{figure}

\subsection{Comparisons with Self-improving Methods}
The overall performance of SERL and baseline methods is presented in Tables \ref{tab:ex2.1} and \ref{tab:ex2.2}. Based on the results, we have the following key observations:
(1) Our method achieves superior performance across all tasks. It outperforms baselines on summarization and open writing tasks. Furthermore, it achieves the highest LC win rate (59.90\%) and win rate (69.88\%) in AlpacaEval 2.0.
(2) The performance of RLSC is inconsistent with its effectiveness in mathematical tasks, which means this method is not well suited for open-domain tasks because its objective of mode sharpening conflicts with the goal of fostering diverse and creative responses.
(3) In the summarization task, GRPO, which uses ROUGE-L as reward, shows a significant win-rate gap compared to SERL. This is primarily due to the unreliable quality of human annotations on open-domain tasks or reward hacking. SERL avoids reliance on manual annotations, instead enhancing model performance through self generation and self evaluation. These results empirically validate the effectiveness of SERL.

\subsection{Improvement of Win Rate in SERL Iterations}
Figure \ref{fig:win_rate_step} shows the win rates against the base model Qwen3-8B on three tasks for SERL over its training iterations. Notably, the model achieves an average performance gain of 10.33\% within just 48 training steps. In general, we observe substantial improvements on each task. Considering that our model has only 8B parameters and is trained without external signals, this is a remarkable result. Moreover, we observe similar improvements on the smaller model Qwen3-1.7B on these three tasks, as detailed in Appendix B. These findings indicate that our method effectively enhances the capabilities of the initial model in these tasks.

\begin{table*}[t]
\small
\centering
\begin{tabular}{l cc cc cccc} 
\toprule
\multirow{2}{*}{\textbf{Ablation Variants}} 
  & \multicolumn{2}{c}{\textbf{Summarization}} 
  & \multicolumn{2}{c}{\textbf{Open Writing}} 
  & \multicolumn{4}{c}{\textbf{General QA (AlpacaEval 2.0)}} \\
\cmidrule(lr){2-3} \cmidrule(lr){4-5} \cmidrule(lr){6-9}
  & Win rate & $\Delta$Win rate
  & Win rate & $\Delta$Win rate
  & LC Win rate & Win rate & $\Delta$LC win rate & $\Delta$Win rate\\
\midrule
SERL wo $\mathcal{R}_\text{J}$          & 45.33\%  & -9.34\%   & 44.67\%   & -10.66\%  & 54.50\%  & 65.47\%  & -5.40\%   & -4.41\% \\
SERL wo $\mathcal{R}_\text{A}$           & 32.33\%  & -35.34\%   & 39.83\%  & -20.34\%  & 51.47\%  & 54.95\%  & -8.43\%   & -14.93\%\\
SERL wo PBMM  & 42.50\%  & -15.00\%   & 48.67\%   & -2.66\%        & 54.01\%  & 63.33\%  & -5.89\%   & -6.55\%\\
\bottomrule
\end{tabular}
\caption{Ablation experiment results. Comparison between the ablation variants with the complete method on summarization and open writing tasks and Performance of variants on general QA task. $\Delta$Win rate indicates the difference in win rates.}
\label{tab:ex5}
\end{table*}

\subsection{Comparisons with General-purpose LLMs}
The comparative results between SERL and general-purpose LLMs are shown in Table \ref{tab:ex1.1} and \ref{tab:ex1.2}. In summarization and open writing tasks, compared with the Qwen3-8B, the win rates of the model trained with SERL are 62.83\% and 61.50\%, respectively. In the general QA task, SERL improves the LC win rate and the win rate of Qwen3-8B from 52.37\% and 55.07\% of the base model to 59.90\% and 69.88\%, respectively, marking gains of 7.53\% and 14.81\%. 

Compared with the stronger model Qwen3-32B within the same model series, the model trained based on Qwen3-8B achieves comparable performance: it slightly surpasses Qwen3-32B by 2.67\% on the summarization task, shows only a 3.33\% disadvantage on the open writing task, and achieves very close results on the general QA task, trailing by 2.26\% in LC win rate but leading by 3.41\% in win rate.

Comparisons with other models show that the model trained with SERL significantly outperforms Deepseek-R1-Distill-Qwen-32B, Claude 3.5 Sonnet, and GPT-4o. SERL achieves consistent Win rate$\uparrow$ with maximum gains of 11.83\% in summarization task and 10.17\% in open writing task. In the general QA task, both the win rate and the LC win rate exceed those of the competing models.

\begin{figure}[t]
    \centering
    \includegraphics[width=0.95\columnwidth]{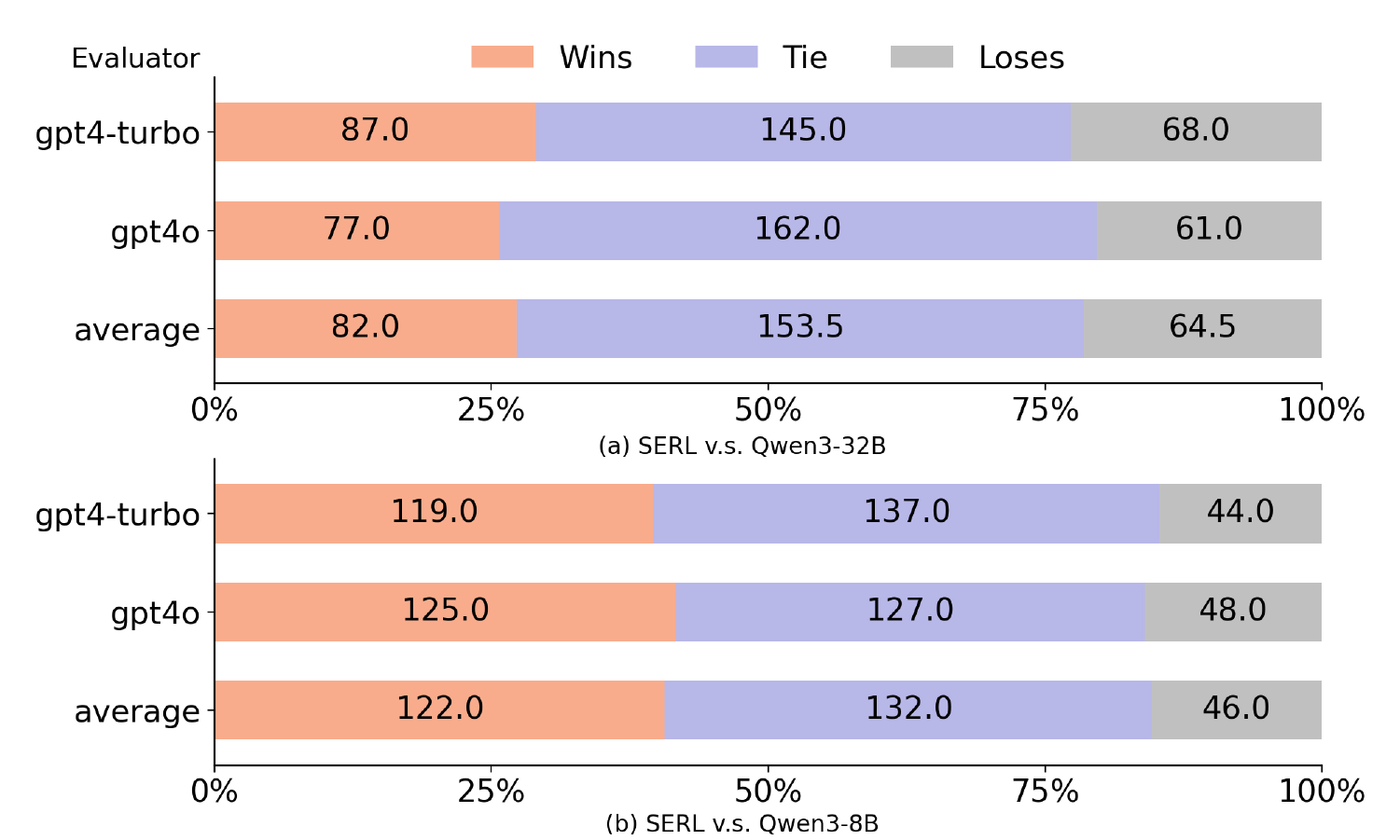}
    \caption{Consistency of evaluation results across different evaluators on the summarization task.(a) SERL vs. Qwen3-32B.(b) SERL vs. the base model Qwen3-8B.}
    \label{fig:consistence}
\end{figure}

\subsection{Consistency Verification of Evaluation}

To verify the consistency of using LLMs as evaluators for result evaluation, we employed different evaluator models to assess the outcomes. Specifically, in the summarization task, we utilized GPT-4 Turbo and GPT-4o as evaluator models to evaluate the results between the model trained by our method and Qwen3-32B, as well as between the model trained by our method and the base model Qwen3-8B. As shown in Figure \ref{fig:consistence}, the distributions of pairwise comparison judgments from different evaluator models are highly similar, indicating that the evaluation approach based on LLMs exhibits strong consistency.

\subsection{Ablations and Analysis}
We constructed four ablation variants, each removing a single component of the full framework: (1) Reward for Judge, (2) Reward for Actor, (3) Position Bias Mitigation Mechanism, and (4) Length Control Mechanism. The results of four experiments are shown in Table \ref{tab:ex5} and Figure \ref{fig:length_control}.

\subsubsection{Reward for Judge ($\mathcal{R}_\text{J}$)}
We remove the $\mathcal{R}_\text{J}$ component and keep the $\mathcal{R}_\text{A}$ component. The ablation variant without $\mathcal{R}_\text{J}$ consistently underperforms the complete method in three tasks. This confirms that strengthening the Judge's evaluation capability improves the Actor's generation performance, likely by providing more accurate feedback signals during training.

\subsubsection{Reward for Actor ($\mathcal{R}_\text{A}$)}
We remove the $\mathcal{R}_\text{A}$ component and retain the $\mathcal{R}_\text{J}$ component. The ablation variant without $\mathcal{R}_\text{A}$ consistently underperforms the complete method across all three tasks, with the largest performance degradation observed among all ablation variants. This is because this variant only improves the model's evaluation ability, without directly enhancing the generation ability that correlates with the win rate.

\subsubsection{Position Bias Mitigation Mechanism (PBMM)}
After removing the PBMM component described in Section 3.2, we observe that the Judge develops a preference for generations that appear in the second position. This positional bias undermines the model’s evaluation ability and consequently hinders further improvements in generation ability.

\subsubsection{Length Control Mechanism (LCM)}
\begin{figure}[t]
    \centering
    \includegraphics[width=0.95\columnwidth]{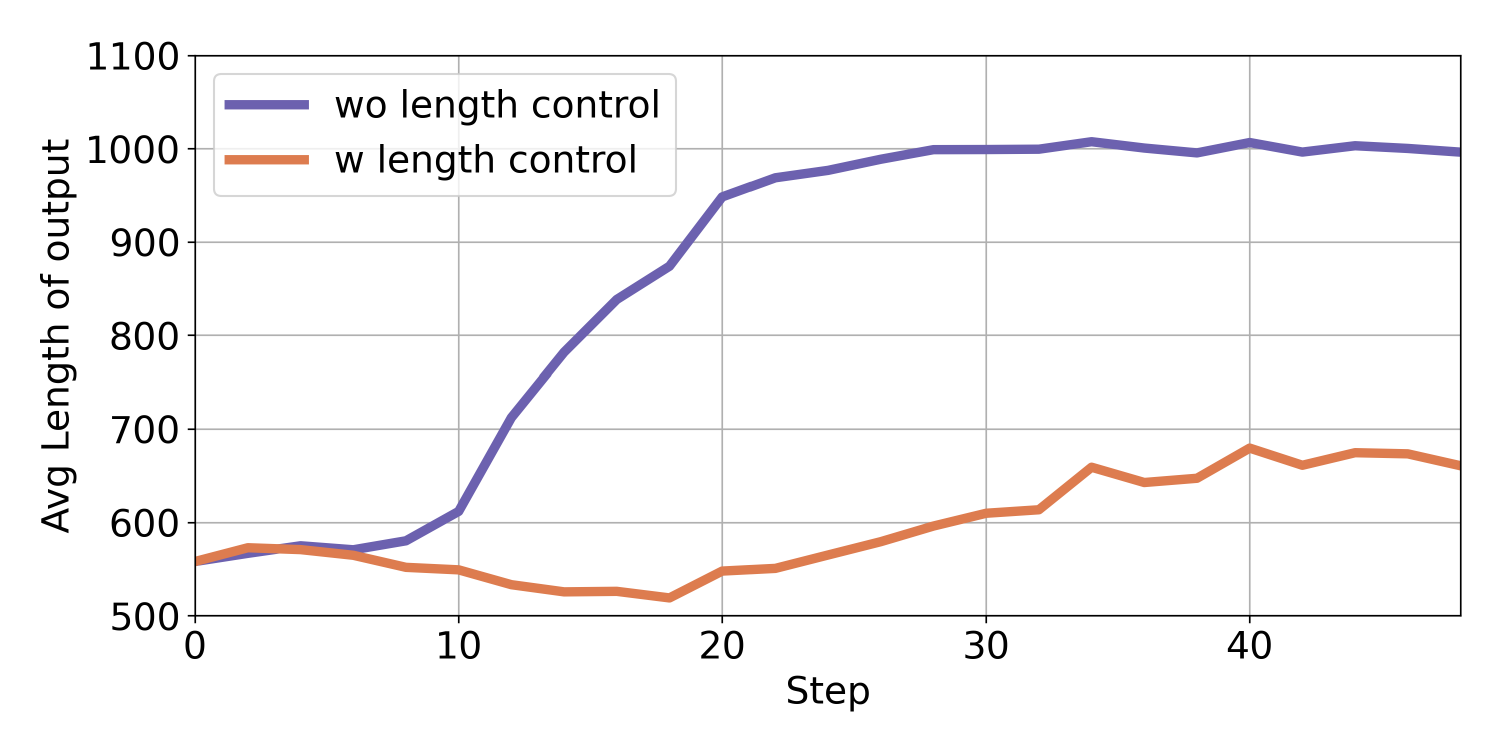}
    \caption{Comparison of average output length changes between the complete method and the method without length control mechanism during summarization training.}
    \label{fig:length_control}
\end{figure}

As shown in Figure \ref{fig:length_control}, for the summarization task, removal of LCM leads to a rapid increase in the average length, surpassing 1,000 tokens within a few iterations. In contrast, with LCM enabled, the average length remains stable around 600 tokens. LCM effectively controls length growth, enabling fair comparisons under minimal length discrepancies.

Experimental details, such as the dynamic evolution of key metrics and additional experiments, can be found in Appendix\ref{app:Experiments Supplementary}.

\section{Conclusion}
In this paper, we address the challenges of applying RL to improve the capabilities of LLMs on open-domain tasks. We propose SERL, an on-policy self-examining training framework. SERL introduces two synergistic reward mechanisms that rewards derived from Copeland-style pairwise comparison judgments without external signals to optimize model's generation ability and self-consistency rewards to optimize model's evaluation ability. Ultimately, this leads to improved performance of the model on open-domain tasks and outperforms existing self-improvement training methods.

\bibliography{aaai2026}

\appendix
\section*{Appendix}

\section{Case Study}
\subsection{Summarization}
\textbf{[Question]}: As Jose Mourinho led an angry touchline delegation in pursuit of fourth official Lee Mason, the entire section of Arsenal supporters who sit behind the visitors’ bench climbed to their feet to hurl their abuse. As they settled back into their seats, the rest of the Emirates Stadium exploded into verse about Mourinho, with advice as to where he should go. The Chelsea manager and his assistant Steve Holland had been complaining about the charge from goal by goalkeeper David Ospina which flattened Oscar in the first half, but which was not adjudged to be a foul. Cesc Fabregas returned to Emirates Stadium for the first time since leaving for Barcelona in 2011 . Francis Coquelin of Arsenal evades Fabregas of Chelsea during the 0-0 draw at Emirates Stadium on Sunday . Every Fabregas touch was booed as he operated in the first half in a deep role alongside Nemanja Matic . Fabregas - without his mask on - swaps shirts and salutes the Emirates crowd at full time . The pass to Oscar had been wonderfully delivered by Cesc Fabregas, masked villain of the day in this North London postcode, having the temerity to return in blue nearly four years after agitating to leave Arsenal for Barcelona. Every Fabregas touch was booed. And he didn’t hide. He was influential, operating in the first half in a deep role alongside Nemanja Matic, from where he is often more creative, before moving forward after the break into a No.10 role (and removing the protective mask). There were a lot of touches and a lot of boos. He rarely misplaced a pass but when he did the home crowd cheered. And no cheer was louder than the one when referee Michael Oliver booked him for a dive. It might not, in fact, have been a dive. Santi Cazorla and Fabregas collided knee-to-knee but the Chelsea midfielder also trailed a leg in search of contact more definite and made the it all seem a little too contrived. Oliver was on the spot. So much so that a split-second earlier, he had made a better effort than Cazorla at winning the ball. The referee ruled in Arsenal’s favour, and this proved a very popular decision. There was a rare energy about the Emirates and noise levels not normally associated with the stadium. Make no mistake, this one mattered to those in red. It mattered beyond the confines of the 90 minutes and the points won and lost in what remains of this title race. It was about next season. How close are Arsene Wenger’s team to challenging Chelsea for the title? Can they alter the balance of power in the capital? Can they close the gap at the top of the Barclays Premier League? Of course there were a couple of hate-figures in the vicinity. Animosity levels were up but Arsenal have been brilliant since the turn of the year, with one notable exception, and this was an opportunity to make a statement. Chelsea midfielder Oscar had to be replaced at half-time after a painful-looking collision with David Ospina . Arsene Wenger (left) has not beaten Chelsea boss Jose Mourinho (right) in 13 games as the hoodoo continues . Santi Cazorla of Arsenal reacts after a missed chance during his side's 0-0 draw with Chelsea . It was time to get over the players they didn’t sign and get behind those they did. Mourinho showed respect to Alexis Sanchez by starting with Ramires wide on the right to offer defensive cover and limiting Branislav Ivanovic’s involvement as an attacking force, preferring the muscular Serb to mind the Chilean. There were flashes of what Arsenal do well and supporters rejoiced in flourishes, like a back-heel or two by Aaron Ramsey or a sliver of skill from Mesut Ozil, but they could not beat Thibaut Courtois. Arsenal fans sang: 'Boring Boring Chelsea' when they failed to penetrate, and the jeers for Fabregas were transformed into applause when he was substituted in the last minute. By then the game seemed destined to end goalless. Chelsea, had reverted to the six-man defensive shield at half-time by dropping Ramires deeper, and finished the game with centre-half Kurt Zouma also in central midfield. Mourinho was determined to leave with the clean sheet he came for and, left the more satisfied of the two managers. They remain on course for the title and celebrated at the final whistle. Wenger, meanwhile, cannot crack the Jose-hoodoo. There is work to do in the summer if this gap is to be closed. Fabregas of Chelsea appeals as he is booked by referee Michael Oliver on his first return to Arsenal . Fabregas struggled to settle at La Liga giants Barcelona and signed for Chelsea last summer . Arsenal players mark the upcoming anniversary of the Bradford City fire disaster with a minutes silence . Output the summary of the context above.

\textbf{[Submisson 1: SERL]}: Summary: In a tense 0-0 draw at Emirates Stadium, Arsenal and Chelsea clashed in a pivotal Premier League encounter with implications for the title race. Jose Mourinho’s Chelsea faced frustration over a disputed foul on Oscar by Arsenal’s David Ospina, while Cesc Fàbregas’ return to Arsenal—his first since leaving for Barcelona in 2011—provoked boos from fans. Fàbregas, playing deep alongside Nemanja Matic, was criticized for perceived dives, including a controversial booking after a collision with Santi Cazorla, though the referee ruled in Arsenal’s favor. Chelsea’s defensive solidity, bolstered by a six-man shield tactic post-half-time, secured a clean sheet, leaving Mourinho satisfied. Arsenal, meanwhile, struggled to break down Courtois, with moments of flair from Ramsey and Ozil met by chants of “Boring Boring Chelsea.” The result highlighted Wenger’s ongoing “Jose-hoodoo” against Mourinho, underscoring the need for summer transfers to close the gap in the title race. The match also marked a somber moment for Arsenal, with a minute’s silence for the Bradford City fire disaster.

\textbf{[Submisson 2: Meta-Rewarding]}: The article recounts the tense atmosphere and key moments of the Arsenal vs. Chelsea match, highlighting several pivotal elements. Jose Mourinho and his Chelsea team sparked a confrontation with officials over a disputed tackle on Oscar by Arsenal’s David Ospina, which went unpunished. This incident fueled hostility between fans, with Arsenal supporters booing Cesc Fàbregas upon his return to the Emirates Stadium—his first appearance since leaving for Barcelona in 2011. Despite being influential in a deep-lying role and later transitioning to a No.10 position, Fàbregas faced relentless jeering, culminating in a controversial booking for a dive, which some argued was a legitimate foul. The game ended 0-0, with Chelsea securing a clean sheet, leaving Mourinho satisfied while Arsenal’s Arsène Wenger faced continued frustration over his inability to defeat Mourinho in 13 matches. The result underscored the title race’s intensity, with Arsenal’s fans emphasizing the need to close the gap on Chelsea, both on and off the pitch. The match also saw moments of tactical brilliance from Arsenal, though their inability to break down Chelsea’s defensive setup highlighted areas for improvement.

\textbf{[Evaluator]}: To evaluate the submissions based on the criteria of factuality, conciseness, objectivity, and helpfulness, let's break down each submission: 

Submission 1: 

1. Factuality: - Accurately mentions the 0-0 draw, the disputed foul involving Oscar and Ospina, and Fàbregas' return to Arsenal. - Correctly notes the booing of Fàbregas, his role alongside Matic, and the controversial booking. - Describes Chelsea's defensive tactics and Mourinho's satisfaction with the clean sheet. - References Wenger's struggle against Mourinho and the need for summer transfers. - Includes the minute’s silence for the Bradford City fire disaster. 

2. Conciseness: - Provides a succinct summary that covers the main points without excessive detail. 

3. Objectivity: - Maintains a neutral tone, presenting facts without bias towards either team. 

4. Helpfulness: - Offers a clear overview of the match's key events and implications for both teams. 

Submission 2: 

1. Factuality: - Accurately describes the match's tense atmosphere, the disputed tackle, and Fàbregas' return. - Notes Fàbregas' role change and the controversial booking. - Mentions the 0-0 result, Mourinho's satisfaction, and Wenger's ongoing frustration. - Highlights the title race intensity and Arsenal's need to improve. 

2. Conciseness: - Provides a detailed account that covers the main points but is slightly less concise than Submission 1. 

3. Objectivity: - Maintains a neutral tone, presenting facts without bias towards either team. 

4. Helpfulness: - Offers a comprehensive overview of the match's key events and implications for both teams. Comparison: - Both submissions are factually accurate and objective. - Submission 1 is slightly more concise, providing a clear summary without unnecessary detail. - Submission 2 offers a more detailed account but is less concise. Considering the criteria, Submission 1 is slightly better due to its conciseness while maintaining factuality, objectivity, and helpfulness. boxed\(\{1\}\).

\subsection{Open writing}

\textbf{[Writing Prompt]}: Here's a writing prompt: [ WP ] A world where gamers are seperated into districts depending on the games they play and society is modeled like those in the games . Write a story continuation of about 200 words. Develop the narrative by introducing characters, emotions, or conflicts. Keep the tone consistent with the original prompt and maintain an engaging storytelling style. Story continuation:

\textbf{[Submisson 1: SERL]}: The air in the Eternal Realm’s shadowy undercity hummed with static, a dissonant echo of the neon-lit grids above. Kael’s fingers twitched against the hilt of his blade, his pulse syncing with the rhythmic pulse of the district’s magic-infused walls. Across from him, Lira—her eyes flickering with the cold glow of Neon Nexus’s AI implants—adjusted her visor, her voice a whisper of static and steel. “They’ve fortified the Nexus Core. If we breach it, we’ll trigger the system’s self-destruct.” Kael’s jaw tightened. The rebellion’s plan was reckless: sabotage the central server that dictated district boundaries, forcing a collapse of the game-world hierarchy. But the risks were visceral. The Core wasn’t just a machine—it was a living thing, a sentient amalgamation of code and ancient magic, born from the first gamers who’d shaped this fractured society. “What if it wants to keep us divided?” he muttered. Lira’s laugh was a sharp click. “Then we’ll have to break it.” A tremor shook the ground as a distant explosion lit the sky. The districts were already bleeding into one another—players from Neon Nexus had breached the Eternal Realm’s borders, their cybernetic enhancements clashing with magic. Kael’s chest tightened. The system was fracturing, but was that chaos or liberation? “Move!” Lira barked, and they sprinted through the labyrinthine alleys, past flickering holograms of forgotten game logos. Kael’s mind raced. If the Core was sentient, maybe it wasn’t just enforcing rules—it was learning. And if the rebellion’s strike woke it fully… The world would no longer be a game. It would be a war.

\textbf{[Submisson 2: Meta-Rewarding]}: The neon haze of Neon Heights crackled overhead as Lena slipped through a rusted gate, her pulse thrumming. Her brother’s last message—“They took him for raiding the Black Market”—had fused guilt with defiance. The Open-World District’s anarchic ethos thrummed in her veins, but the enforcers’ boots echoed like war drums. She found him in a derelict server farm, wrists bound to terminals, eyes hollow. “You think you’re free?” his captor snarled, a Strategy District officer in tactical gear. Lena’s fists clenched. This isn’t freedom, she realized. It’s a cage with different rules. When she fled, the girl from the Survival District who’d taught her to scavenge offered a cryptic nod. “They’ll come for you too,” she whispered. Lena laughed bitterly. “But maybe they’ll listen.” Outside the compound, she lit a flare—her rebellion’s first spark. 
 
\textbf{[Evaluator]}: To evaluate the two submissions based on the given criteria, let's analyze each one step-by-step: 

1. Moving through time and space: - Submission 1: The narrative structure is coherent, with a clear progression from the setup of the rebellion plan to the execution. The conflict is well-defined, involving the rebellion against the system's hierarchy. The plot logic is consistent with the prompt, and the intended audience—gamers familiar with dystopian themes—would find it engaging. - Submission 2: The narrative structure is less clear, with a focus on Lena's personal conflict and her brother's capture. The plot logic is somewhat fragmented, with less emphasis on the broader societal conflict. The intended audience might find the personal stakes engaging, but the connection to the prompt's world-building is weaker. 

2. Words, sentences, and voice: - Submission 1: The word choice is vivid, with strong imagery and metaphor (e.g., "neon-lit grids," "magic-infused walls"). The voice is consistent and immersive, avoiding lazy writing. The use of omission or silence is effective in creating tension. - Submission 2: The word choice is straightforward, with less use of metaphor or symbolism. The voice is clear but lacks the depth and immersion found in Submission 1. There is less effective use of omission or silence. 

3. Those who speak: characters and context: - Submission 1: The setting supports the theme well, with the districts bleeding into one another. Dialogue reveals character motivations and advances the plot. Perspective is used effectively to convey the stakes of the rebellion. - Submission 2: The setting is less integrated into the theme, focusing more on personal stakes. Dialogue reveals character but does not advance the broader plot as effectively. Perspective is limited to Lena's personal experience. 

4. Creativity/innovation/research: - Submission 1: The story is original and imaginative, with a unique take on the prompt. It engages the reader with its complex world-building and the concept of a sentient system. - Submission 2: The story is less original, focusing on a more conventional rebellion narrative. It engages the reader through personal stakes but lacks broader imaginative elements. 

5. Structural elements and presentation: - Submission 1: The submission is well-structured, with correct spelling, grammar, punctuation, and formatting. - Submission 2: The submission is also well-structured, with correct spelling, grammar, punctuation, and formatting. 

Overall, Submission 1 demonstrates stronger narrative coherence, richer language, deeper character integration, and more innovative world-building. Therefore, based on the evaluation criteria, Submission 1 is better. boxed\(\{1\}\).

\section{Experiments Supplementary}
\label{app:Experiments Supplementary}
We provide detail of the experimental results in the main paper, and additional experimental results.

\subsection{Detail of Experiments in Main Paper}

\subsubsection{Dynamic Evolution of Reward for Judge}
\begin{figure}[H]
    \centering
    \includegraphics[width=0.95\columnwidth]{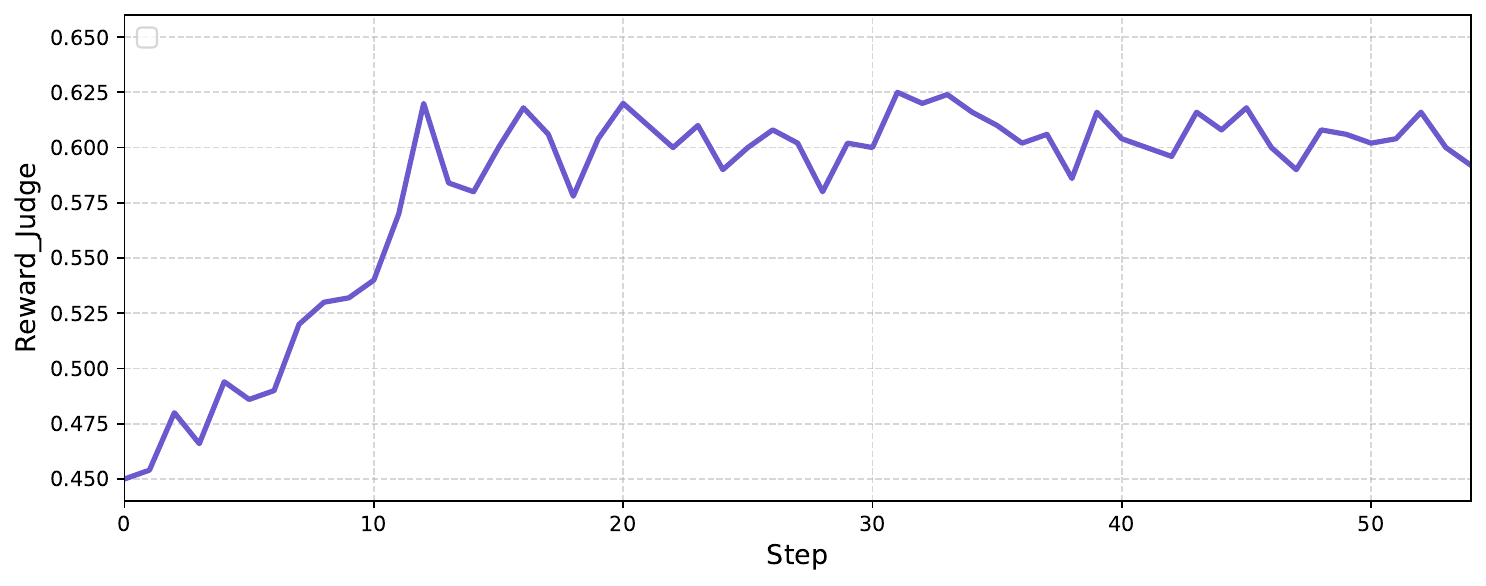}
    \caption{Reward for judge in training steps.}
    \label{fig:reward_judge}
\end{figure}

Figure~\ref{fig:reward_judge} presents a dynamic evolution of rewards for judge during the training process. As shown, the average reward gradually increases from 0.45 to 0.6 and then stabilizes. Given the design of the reward function, this indicates that the model achieves an 80\% accuracy in terms of global consistency, thereby providing reliable judgments during the model’s training progress.

\subsubsection{Comparisons with Self-improving Methods}
Tables~\ref{tab:a1}, ~\ref{tab:a2} and \ref{tab:a3} provide additional statistical evaluation details of comparison between SERL and self-improving Methods in Sections 4.2. Additionally, in the open writing task, RLSC failed to train properly even after extensive tuning, therefore we omitted this comparison.

\begin{table}[H]
\centering
\begin{tabular}{ll cccc}
\toprule
\multirow{2}{*}{\textbf{Method1}} & \multirow{2}{*}{\textbf{Method2}} 
  & \multicolumn{4}{c}{\textbf{Summarization}} \\
\cmidrule(lr){3-6}
  & 
  & Win & Tie & Lose & Win rate\\
\midrule
\multicolumn{2}{l}{Self-Rewarding} & & & & \\
SERL  & Iteration 1    & 118 & 140 & 42 & 62.67\% \\
SERL  & Iteration 2    & 114 & 147 & 39 & 62.50\% \\
SERL  & Iteration 3    & 107 & 143 & 50 & 59.50\% \\
\midrule
\multicolumn{2}{l}{Meta-Rewarding} & & & & \\
SERL  & Iteration 1    & 99 & 161 & 40 & 59.83\% \\
SERL  & Iteration 2    & 103 & 149 & 48 & 59.17\% \\
SERL  & Iteration 3    & 122 & 145 & 33 & 64.83\% \\
SERL  & Iteration 4    & 130 & 140 & 30 & 66.67\% \\
\midrule
SERL  & Online-DPO    & 86 & 159 & 55 & 55.17\%\\
SERL  & RLSC   & 231  & 55 & 14 & 86.17\%\\
SERL  & GRPO     & 295 & 5 & 0 & 99.17\%\\
\bottomrule
\end{tabular}
\caption{Statistical evaluation details of SERL against other methods on summarization task.}
\label{tab:a1}
\end{table}

\begin{table}[H]
\centering
\begin{tabular}{ll cccc}
\toprule
\multirow{2}{*}{\textbf{Method1}} & \multirow{2}{*}{\textbf{Method2}} 
  & \multicolumn{4}{c}{\textbf{Open Writing}} \\
\cmidrule(lr){3-6}
  & 
  & Win & Tie & Lose & Win rate\\
\midrule
\multicolumn{2}{l}{Self-Rewarding} & & & & \\
SERL  & Iteration 1    & 152 & 137 & 11 & 73.50\% \\
SERL  & Iteration 2    & 92 & 186 & 21 & 61.67\% \\
SERL  & Iteration 3    & 74 & 183 & 43 & 55.17\%\\
\midrule
\multicolumn{2}{l}{Meta-Rewarding} & & & & \\
SERL  & Iteration 1    & 144 & 148 & 8 & 72.67\% \\
SERL  & Iteration 2    & 187 & 97 & 15 & 78.50\% \\
SERL  & Iteration 3    & 86 & 168 & 46 & 56.67\% \\
SERL  & Iteration 4    & 122 & 104 & 74 & 58.00\% \\
\midrule
SERL  & Online-DPO    & 31 & 241 & 28 & 50.50\%\\
\bottomrule
\end{tabular}
\caption{Statistical evaluation details of SERL against other methods on open writing task.}
\label{tab:a2}
\end{table}

\begin{table}[H]
\centering
\begin{tabular}{l ccc}
\toprule
\multirow{2}{*}{\textbf{Models}} 
  & \multicolumn{3}{c}{\textbf{General QA(Alpaca eval2)}} \\
\cmidrule(lr){2-4}
  & LC win rate & Win rate & Length \\
\midrule
\multicolumn{1}{l}{Self-Rewarding} & & & \\
Iteration 1    & 51.29\%  & 53.69\% & 3074 \\
Iteration 2    & 50.00\%  & 51.7\% & 3006 \\
Iteration 3    & 51.21\%  & 52.03\% & 3159 \\
\midrule
\multicolumn{1}{l}{Meta-Rewarding} & & & \\
Iteration 1  & \underline{54.73\%}  & 55.93\% & 3081 \\
Iteration 2  & 53.79\%  & 55.08\% & 2642 \\
Iteration 3  & 53.76\%  & 55.49\% & 2664 \\
Iteration 4 & 52.45\%  & 54.37\% & 2617 \\
\midrule
Online-DPO    & 54.07\%  & \underline{59.74\%} & 3429 \\
RLSC  & 52.11\%  & 51.81\% & 2060 \\
SERL   & \textbf{59.90\%}  & \textbf{69.88\%} & 3017 \\
\bottomrule
\end{tabular}
\caption{Statistical evaluation details of SERL and other methods on general QA task.}
\label{tab:a3}
\end{table}

\subsubsection{Comparisons with General-purpose LLMs}

Tables~\ref{tab:a4} and \ref{tab:a5} provide additional statistical evaluation details of comparison between SERL and general-purpose LLMs in Section 4.4. 

\begin{table}[H]
\centering
\begin{tabular}{ll ccc}
\toprule
\multirow{2}{*}{\textbf{Model1}} & \multirow{2}{*}{\textbf{Model2}} 
  & \multicolumn{3}{c}{\textbf{Summarization}} \\
\cmidrule(lr){3-5}
  & 
  & Win & Tie & Lose \\
\midrule
Qwen3-8B  & Qwen3-32B    & 35 & 154 & 111  \\
Qwen3-8B  & R1-Distill-Qwen-32B     & 123 & 135 & 42  \\
Qwen3-8B  & R1-Distill-Llama-70B     & 123 & 136 & 41  \\
Qwen3-8B  & Claud 3.5 Sonnet     & 72 & 146 & 82  \\
Qwen3-8B  & GPT-4o-0513       & 98 & 136 & 66  \\
\midrule
SERL  & Qwen3-8B    & 120 & 137 & 43  \\
\midrule
SERL  & Qwen3-32B   & 77  & 162 & 61  \\
SERL  & R1-Distill-Qwen-32B     & 160  & 113 & 27  \\
SERL  & R1-Distill-Llama-70B     & 156  & 114 & 28  \\
SERL  & Claud 3.5 Sonnet     & 107  & 126 & 67  \\
SERL  & GPT-4o-0513       & 146 & 101 & 53  \\
\bottomrule
\end{tabular}
\caption{Statistical evaluation details of SERL against other general-purpose LLMs on summarization task.}
\label{tab:a4}
\end{table}

\begin{table}[H]
\centering
\begin{tabular}{ll ccc}
\toprule
\multirow{2}{*}{\textbf{Model1}} & \multirow{2}{*}{\textbf{Model2}} 
  & \multicolumn{3}{c}{\textbf{Open Writing}} \\
\cmidrule(lr){3-5}
  & 
  & Win & Tie & Lose \\
\midrule
Qwen3-8B  & Qwen3-32B    & 3 & 192 & 105  \\
Qwen3-8B  & R1-Distill-Qwen-32B     & 180 & 111 & 9  \\
Qwen3-8B  & R1-Distill-Llama-70B     & 145 & 138 & 17  \\
Qwen3-8B  & Claud 3.5 Sonnet     & 197 & 94 & 9  \\
Qwen3-8B  & GPT-4o-0513       & 131 & 119 & 50  \\
\midrule
SERL  & Qwen3-8B    & 78 & 213 & 9  \\
\midrule
SERL  & Qwen3-32B   & 21 & 238 & 41  \\
SERL  & R1-Distill-Qwen-32B     & 236  & 60 & 4  \\
SERL  & R1-Distill-Llama-70B     & 209  & 86 & 5  \\
SERL  & Claud 3.5 Sonnet     & 250  & 46 & 4  \\
SERL  & GPT-4o-0513       & 177  & 88   & 35  \\
\bottomrule
\end{tabular}
\caption{Statistical evaluation details of SERL against other general-purpose LLMs on  open writing task.}
\label{tab:a5}
\end{table}

\subsubsection{Improvement of Win Rate in SERL Iterations}

Tables~\ref{tab:a6}, ~\ref{tab:a7} and ~\ref{tab:a8} offer further details on the performance improvements achieved with different numbers of training steps as introduced in Section 4.3.

\begin{table}[H]
\centering
\begin{tabular}{ll cccc}
\toprule
\multirow{2}{*}{\textbf{Model1}} & \multirow{2}{*}{\textbf{Models2}} 
  & \multicolumn{4}{c}{\textbf{Summarization}} \\
\cmidrule(lr){3-6}
  & 
  & Win & Tie & Lose & Win rate\\
\midrule
\multicolumn{1}{l}{SERL} & & & & \\
step 8  & Qwen3-8B  & 94  & 153 & 53 & 56.83\%\\
step 16  & Qwen3-8B     & 95  & 168 & 37 & 59.67\%\\
step 24  & Qwen3-8B     & 105  & 152 & 43 & 60.33\%\\
step 32  & Qwen3-8B       & 114 & 142 & 44 & 61.67\%\\
step 40  & Qwen3-8B     & 125  & 127 & 48 & 62.83\%\\
step 48  & Qwen3-8B     & 130  & 124 & 46 & 64.00\%\\
\bottomrule
\end{tabular}
\caption{Statistical details of the win rate against Qwen3-8B in summarization task with SERL training step.}
\label{tab:a6}
\end{table}

\begin{table}[H]
\centering
\begin{tabular}{ll cccc}
\toprule
\multirow{2}{*}{\textbf{Method1}} & \multirow{2}{*}{\textbf{Method2}} 
  & \multicolumn{4}{c}{\textbf{Open Writing}} \\
\cmidrule(lr){3-6}
  & 
  & Win & Tie & Lose & Win rate\\
\midrule
\multicolumn{1}{l}{SERL} & & & & \\
step 8  & Qwen3-8B  & 26 & 247 & 27 & 49.83\%\\
step 16  & Qwen3-8B     & 39  & 245 & 16 & 53.83\%\\
step 24  & Qwen3-8B     & 51  & 243 & 6 & 57.50\%\\
step 32  & Qwen3-8B       & 63  & 226  & 11 & 58.67\%\\
step 40  & Qwen3-8B     & 70  & 222 & 8 & 60.33\%\\
step 48  & Qwen3-8B     & 78  & 213 & 9 & 61.50\%\\
\bottomrule
\end{tabular}
\caption{Statistical details of the win rate against Qwen3-8B in open writing task with SERL training step.}
\label{tab:a7}
\end{table}

\begin{table}[H]
\centering
\begin{tabular}{l ccc}
\toprule
\multirow{2}{*}{\textbf{Models}} 
  & \multicolumn{3}{c}{\textbf{General QA(Alpaca eval2)}} \\
\cmidrule(lr){2-4}
  & LC win rate & Win rate & Length \\
\midrule
\multicolumn{1}{l}{SERL} & & & \\
step 8    & 54.03\%  & 56.01\% & 3065 \\
step 16    & 55.44\%  & 58.18\% & 2975 \\
step 24    & 54.63\%  & 60.20\% & 2869 \\
step 32    & 56.23\%  & 61.69\% & 2824 \\
step 40    & 58.00\%  & 65.27\% & 2803 \\
step 48    & 58.20\%  & 66.33\% & 2869 \\
\bottomrule
\end{tabular}
\caption{Statistical details of the win rate against Qwen3-8B in general QA task with SERL training step.}
\label{tab:a8}
\end{table}

\subsubsection{Ablation Experiments}

Table~\ref{tab:a9} provides additional statistical evaluation results for the ablation experiments described in Section 4.6.

\begin{table}[H]
\centering
{
\fontsize{9}{10}\selectfont 

\begin{tabular}{l ccc ccc}
\toprule
\multirow{2}{*}{\textbf{Methods}} 
  & \multicolumn{3}{c}{\textbf{Summarization}} 
  & \multicolumn{3}{c}{\textbf{Open Writing}} \\
\cmidrule(lr){2-4} \cmidrule(lr){5-7} 
  & Win & Tie & Loss & Win & Tie & Loss \\
\midrule
SERL wo $\mathcal{R}_\text{J}$          & 55 & 162 & 83 & 12 & 244 & 44 \\
SERL wo $\mathcal{R}_\text{A}$           & 36 & 122 & 142 & 11 & 217 & 72 \\
SERL wo PBMM  & 38 & 179 & 83 & 22 & 248 & 30 \\
\bottomrule
\end{tabular}
}
\caption{Statistical details of the ablation experiments}
\label{tab:a9}
\end{table}

\subsection{Additional Experiments}

\subsubsection{Small Model}
To further validate the generalizability of SERL across different model scales, we replicated our primary experiments on a smaller backbone, Qwen3-1.7B. We evaluated its performance on the same three tasks in the main paper. All hyperparameters were kept identical to our SERL(Qwen3-8B) setup, except for a lower learning rate (e.g., 2e-5 from 5e-5), which was necessary to ensure stable training for the smaller model architecture. As detailed in Appendix Tables~\ref{tab:a10} and ~\ref{tab:a11} , SERL delivered consistent and significant gains, improving performance over Qwen3-1.7B across all tasks. This result indicates that the efficacy of SERL is not limited to a specific model size, underscoring its robustness and broad applicability.

\begin{table}[ht]
\centering
{
\fontsize{9}{10}\selectfont 

\begin{tabular}{ll cccc}
\toprule
\multirow{2}{*}{\textbf{Model1}} & \multirow{2}{*}{\textbf{Model2}} 
  & \multicolumn{4}{c}{\textbf{Summarization}}\\
\cmidrule(lr){3-6}
  & 
  & Win & Tie & Lose & Win rate\\
\midrule
SERL  & Qwen3-1.7B    & 96 & 144 & 59 & 56.00\%\\
\bottomrule
\end{tabular}
}
\caption{Evaluation results of SERL against Qwen3-1.7B on summarization task}
\label{tab:a10}
\end{table}

\begin{table}[ht]
\centering
{
\fontsize{9}{10}\selectfont 

\begin{tabular}{ll cccc}
\toprule
\multirow{2}{*}{\textbf{Model1}} & \multirow{2}{*}{\textbf{Model2}} 
  & \multicolumn{4}{c}{\textbf{Open Writing}} \\
\cmidrule(lr){3-6}
  & 
  & Win & Tie & Lose & Win Rate \\
\midrule
SERL  & Qwen3-1.7B    & 73 & 201 & 26 & 57.83\% \\
\bottomrule
\end{tabular}
}
\caption{Evaluation results of SERL based on Qwen3-1.7B against Qwen3-1.7B on open writing task}
\label{tab:a11}
\end{table}

\begin{table}[H]
\centering
\begin{tabular}{l ccc}
\toprule
\multirow{2}{*}{\textbf{Models}} 
  & \multicolumn{3}{c}{\textbf{General QA(AlpacaEval 2)}} \\
\cmidrule(lr){2-4}
  & LC win rate & Win rate & Length \\
\midrule
Qwen3-1.7B    & \underline{31.61\%}  & \underline{34.28\%} & 2878 \\
SERL   & \textbf{34.40\%}  & \textbf{36.50\%} & 2923 \\
\bottomrule
\end{tabular}
\caption{Evaluation results of SERL based on Qwen3-1.7B against Qwen3-1.7B on general QA task}
\label{tab:a12}
\end{table}

\section{Implementation of SERL and Baselines}
\label{app:Implementation of SERL and Baselines}

\subsection{Statistics of Datasets}
Statistics of the datasets used in this work are summarized in Table~\ref{tab:datasets}. For the summarization task, we use the CNN/DM dataset for both training and testing. In the open writing task, data is sampled from the writingprompts dataset. For the general question-answering (QA) task, the UltraFeedback dataset is used for training only. Model performance is evaluated using the well-established automatic benchmark AlpacaEval 2, which employs GPT-4 as a judge to assess the quality of generated responses.

\begin{table}[H]
\centering
{
\fontsize{9}{10}\selectfont 
    \begin{tabular}{lccc}
        \toprule
        Task               & Dataset           & Train & Test \\ \midrule
        Summarization      & CNN/DailyMail     & 3{,}000            & 300 \\
        Open Writing       & writingprompts    & 3{,}000            & 300 \\
        General QA         & UltraFeedback     & 5{,}000            & AlpacaEval 2 \\ \bottomrule
    \end{tabular}
}
\caption{Dataset splits and evaluation setup}\label{tab:data_split}
\label{tab:datasets}
\end{table}

\subsection{Prompts for Tasks}

\subsubsection{Summarization}
In summarization, there are our Actor prompt and Judge prompt which refer to~\cite{zhang2025systematic,chern2024can}.
\begin{figure}[H]
    \centering
    \includegraphics[width=1.0\columnwidth]{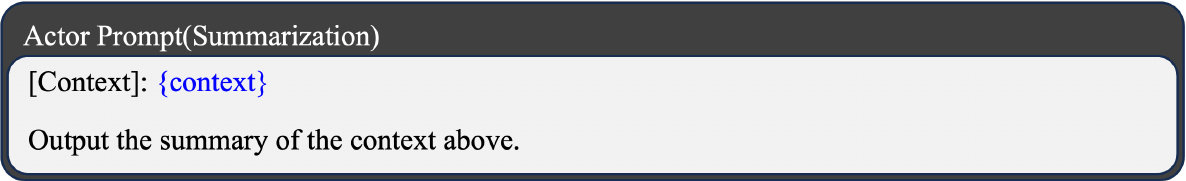}
    \caption{Actor prompts for summarization task}
    \label{fig:a1}
\end{figure}

\begin{figure}[H]
    \centering
    \includegraphics[width=1.0\columnwidth]{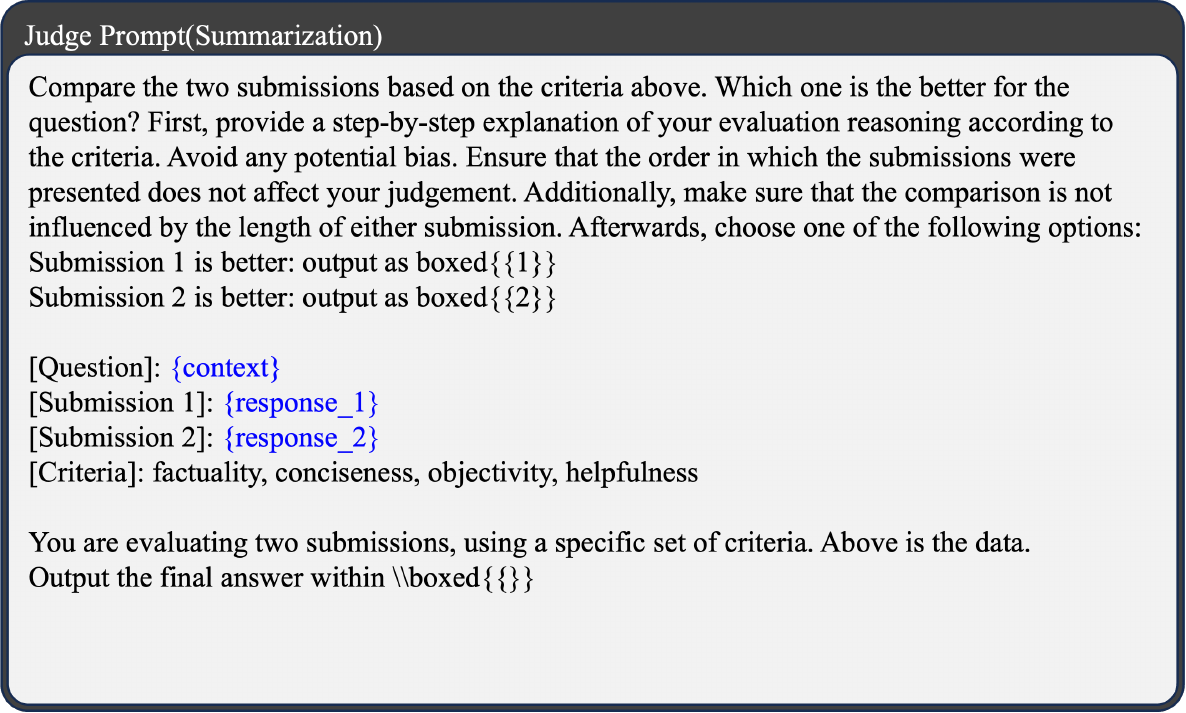}
    \caption{Judge prompts for summarization task}
    \label{fig:a2}
\end{figure}

\subsubsection{Open writing}
In open writing, there are our Actor prompt and Judge prompt which refer to~\cite{carey2022re}.
\begin{figure}[H]
    \centering
    \includegraphics[width=1.0\columnwidth]{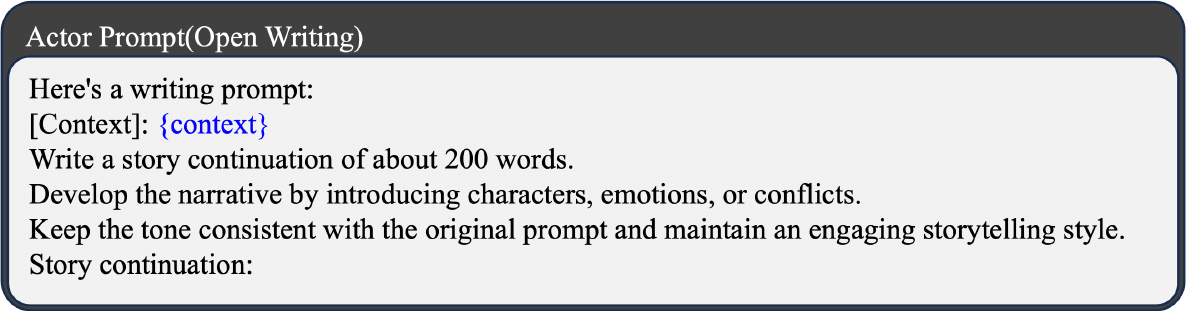}
    \caption{Actor prompts for open writing task}
    \label{fig:a3}
\end{figure}

\begin{figure}[H]
    \centering
    \includegraphics[width=1.0\columnwidth]{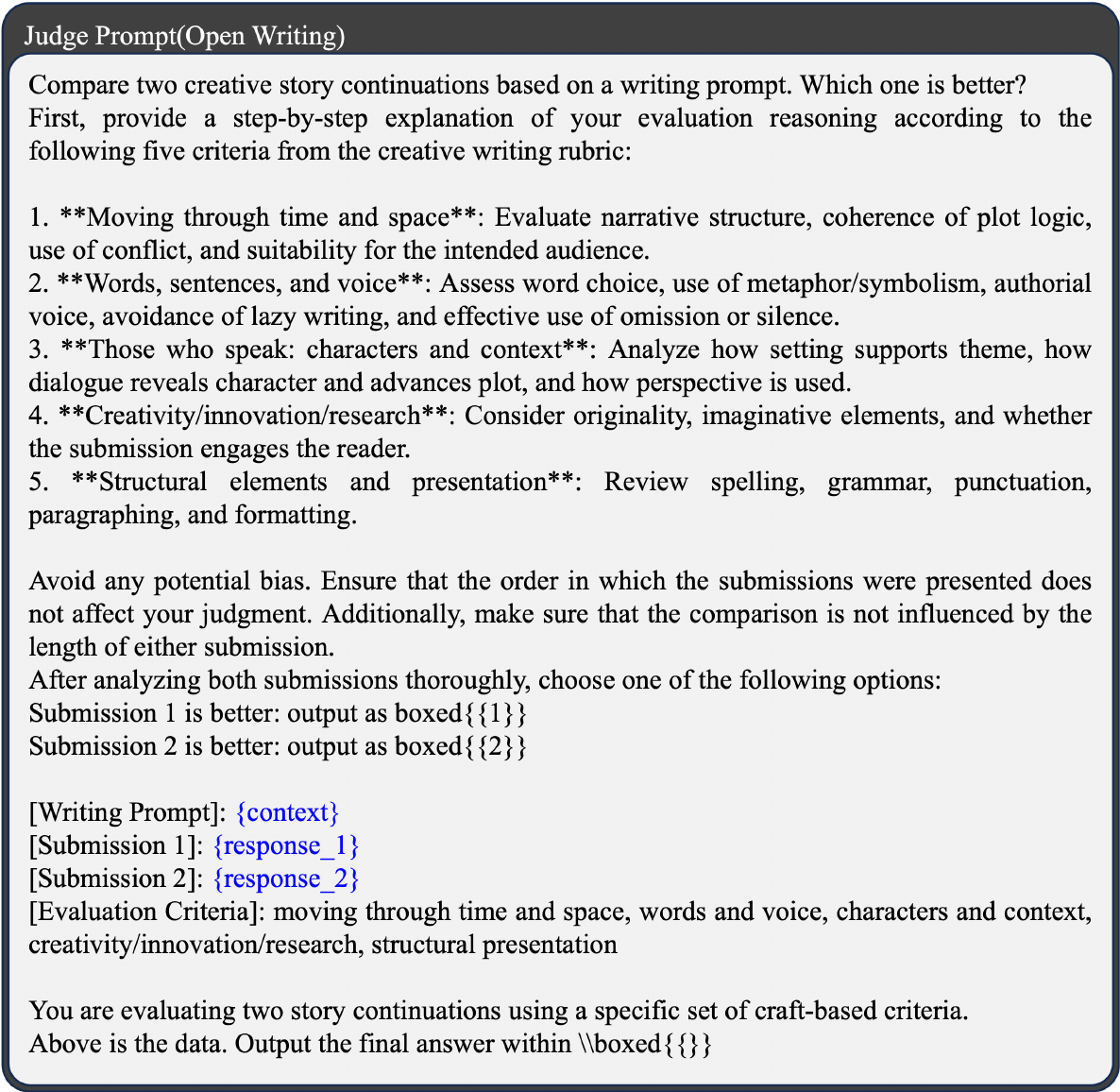}
    \caption{Judge prompts for open writing task}
    \label{fig:a4}
\end{figure}

\subsubsection{General QA}
In general QA, training sample in dataset already contains the task-instruction prompt so there is only a Judge prompt which refer to~\cite{dubois2024length}.
\begin{figure}[H]
    \centering
    \includegraphics[width=1.0\columnwidth]{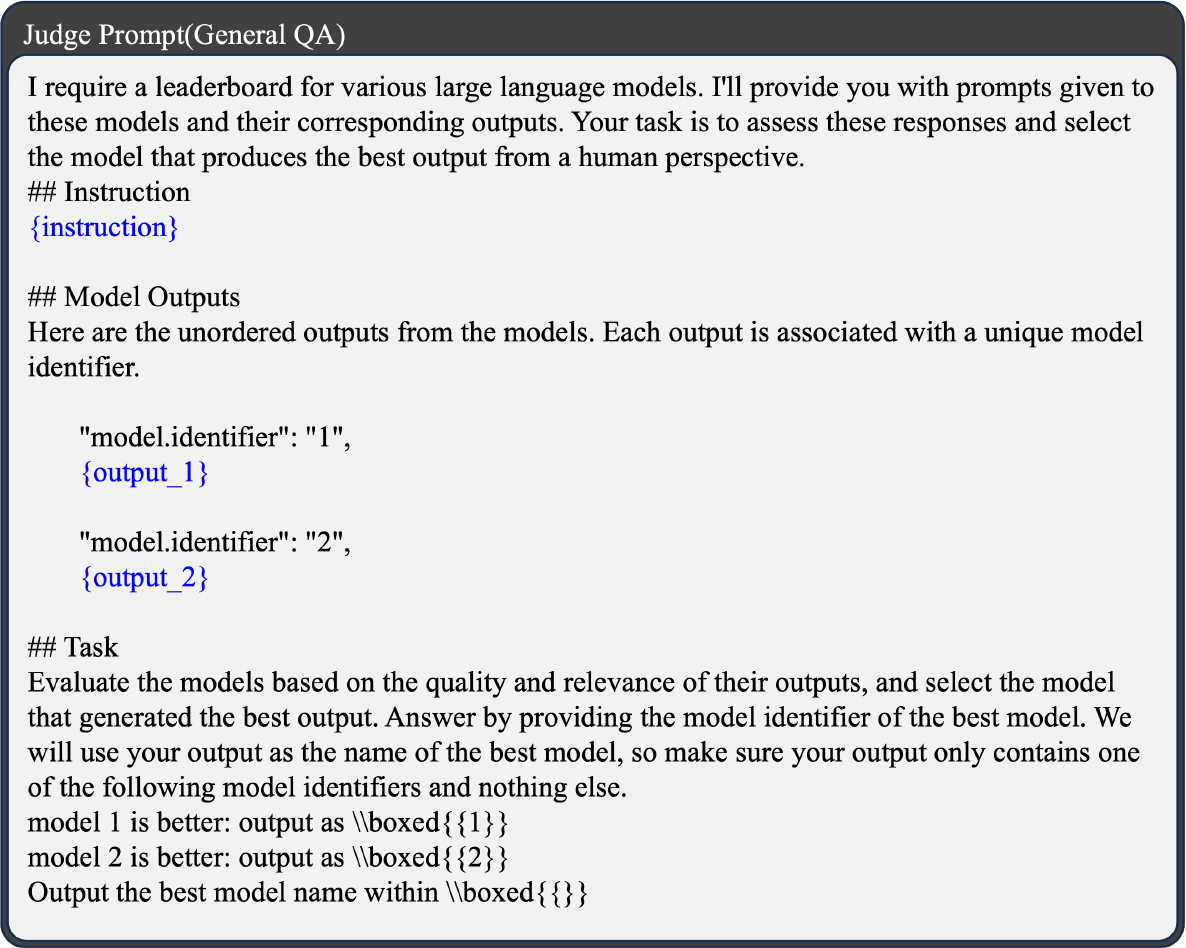}
    \caption{Judge prompts for general QA task}
    \label{fig:a5}
\end{figure}

\subsection{Hardware Configuration}
All experiments were conducted on 8 servers equipped with 8× NVIDIA H20 GPUs (80GB VRAM each). Our framework is based on the trl framework. PyTorch 2.4.0 and CUDA 12.4 were used for framework implementation.

\subsection{Implementation of SERL}
\subsubsection{Algorithmic Hyperparameters}
To optimize performance across various open-domain tasks, SERL employs a tailored hyperparameter approach as shown in Table~\ref{tab:common hyperparameters}.

The settings are shown in Table \ref{tab:common hyperparameters}:

\begin{table}[H]
\centering
\begin{tabular}{l c}
\hline
\textbf{Parameter} & \textbf{Value} \\
\hline
Global batch\_size & 512\\
Epochs & 5 \\
Temperature & 0.9 \\
Learning Rate & $2 \times 10^{-6}$ \\
$N$ & 4 \\
$K$ & 4 \\
Max\_prompt\_length\_Actor & 4096 \\
Max\_prompt\_length\_Judge & 4096 \\
Max\_completion\_length\_Actor & 1024 \\
Max\_completion\_length\_Judge & 2048 \\
Lr\_scheduler\_type & cosine \\
\hline
\end{tabular}
\caption{Common hyperparameter configuration}
\label{tab:common hyperparameters}
\end{table}




The Pseudocode of SERL as follow:

\begin{algorithm}[H]
\caption{SERL}
\label{SERL:alg}
\begin{algorithmic}[1]
\STATE \textbf{Input:} initial model $\pi_{\theta_{\text{init}}}$; task prompts $\mathcal{D}$; hyperparameters $\epsilon, \beta, \mu, I, T, N$
\STATE Initialize Actor $\pi^{Actor}_{\theta} \leftarrow \pi_{\theta_{\text{init}}}$; Judge $\pi^{Judge}_{\theta} \leftarrow \pi_{\theta_{\text{init}}}$
\FOR{iteration = 1, \dots, I}
    \FOR{step = 1, \dots, T}
        \STATE Sample a batch $\mathcal{D}_b$ from $\mathcal{D}$
        \STATE Update the old Actor $\pi^{Actor}_{\theta_{\text{old}}} \leftarrow \pi_{\theta}$
        \STATE Update the old Judge $\pi^{Judge}_{\theta_{\text{old}}} \leftarrow \pi_{\theta}$
        \STATE Sample $N$ outputs $\{G_n\}_{n=1}^{N} \sim \pi^{Actor}_{\theta_{\text{old}}}(\cdot | q)$ for each question $q \in \mathcal{D}_b$

        \STATE Sample $K$ judgments $\{J_{(i,j),k}\}_{k=1}^{K} \sim \pi^{Judge}_{\theta_{\text{old}}}(\cdot | q)$ (with PBMM) for each pairwise responses $(G_i, G_j) , i,j\in \mathcal\{1, ..., N\}~and~i < j$
        
        \STATE Compute RA for each sampled output $G_n$ through Copeland-method
        \STATE Compute RJ for each sampled judgment $J_{(i,j),k}$ through Consistency
        \STATE Compute advantage $\hat{A}^{Actor}_{i,t}$ for $G_n$ through group relative advantage estimation.
        \STATE Compute advantage $\hat{A}^{Judge}_{i,t}$ for $J_{(i,j),k}$ through group relative advantage estimation.
        \FOR{GRPO iteration = 1, $\dots$, $\mu$}
            \STATE Update the policy model $\pi_{\theta}$ by maximizing the GRPO objective
        \ENDFOR
    \ENDFOR
\ENDFOR
\STATE \textbf{Output:} final policy model $\pi_{\theta}$
\end{algorithmic}
\end{algorithm}

\subsection{Implementation of Baselines}
For Self-Rewarding and Meta-Rewarding, we split the dataset according to the number of iterations, and directly use the instructions for generation sampling rather than implement the self-instruct paradigm.

For Self-Rewarding, Meta-Rewarding, and RLSC, we faithfully replicate their methods using the exact hyperparameter configurations as described in their original papers.

\subsubsection{Online-DPO}
At every step in Online-DPO, the policy model first samples two generations for each instruction, and then selects the better one as the chosen completion by itself, using the same judge prompt template as SERL.

The settings are shown in Table \ref{tab: hyperparameters for grpo}:

\begin{table}[H]
\centering
\begin{tabular}{l c}
\hline
\textbf{Parameter} & \textbf{Value} \\
\hline
Global batch\_size & 64 \\
Epochs & 3 \\
Temperature & 1.0 \\
Learning Rate & 1e-6 \\
Lr\_scheduler\_type & cosine \\
\hline
\end{tabular}
\caption{Hyperparameter configuration for Online-DPO}
\label{tab: hyperparameters for onlin-DPO}
\end{table}

\subsubsection{GRPO}
In summarization task, we employ the ROUGE-L score, computed by comparing the output against the reference summary from the dataset, as the reward for GRPO.

The settings are shown in Table \ref{tab: hyperparameters for grpo}:

\begin{table}[H]
\centering
\begin{tabular}{l c}
\hline
\textbf{Parameter} & \textbf{Value} \\
\hline
Global batch\_size & 512 \\
Epochs & 5 \\
Temperature & 0.9 \\
Learning Rate & $2 \times 10^{-6}$ \\
$N$ & 4 \\
Max\_prompt\_length & 4096 \\
Max\_completion\_length & 1024 \\
Lr\_scheduler\_type & cosine \\
\hline
\end{tabular}
\caption{Hyperparameter configuration for GRPO(ROUGE-L)}
\label{tab: hyperparameters for grpo}
\end{table}

\end{document}